# Artificial Intelligence BlockCloud (AIBC) Technical Whitepaper

Prof. Qi Deng
Founder, CEO and Chief Scientist of Cofintelligence Financial Technology Ltd.
qi.deng@cofintelligence.ai

Abstract

The AIBC is an Artificial Intelligence and blockchain technology based large-scale decentralized ecosystem that allows system-wide low-cost sharing of computing and storage resources. The AIBC consists of four layers: a fundamental layer, a resource layer, an application layer, and an ecosystem layer.

The AIBC implements a two-consensus scheme to enforce upper-layer economic policies and achieve fundamental layer performance and robustness: the DPoEV incentive consensus on the application and resource layers, and the DABFT distributed consensus on the fundamental layer. The DABFT uses deep learning techniques to predict and select the most suitable BFT algorithm in order to achieve the best balance of performance, robustness, and security. The DPoEV uses the knowledge map algorithm to accurately assess the economic value of digital assets.

The AIBC is task-driven with a “blocks track task” dynamic sharding structure. It is a 2D BlockCloud with side chains originated from the super nodes that track the progress of tasks. The 2D implementation makes it extremely efficient to evaluate the incremental economic value of additional knowledge contributed by each task. The dynamic sharding feature resolves the scalability issue and improves the efficiency further.

The AIBC has a dual-token implementation. In addition to the system-wide unified measure of value and transaction medium CFTX, each DSOL will have a separate numbering interval as a single distinguishable token, DSOLxxxx. The dual-token approach allows the CFTX be used for the entire AIBC ecosystem, while enables the transfer of DSOL ownership through auctions of DSOLxxxx tokens.

# 1 Introduction

## 1.1 Artificial Intelligence Blockchain (AIBC)

The AIBC is an Artificial Intelligence (AI) based blockchain ecosystem. Anchored on the principles of decentralization, scalability, and controllable cost. Based on the principles of decentralization, scalability and controllable cost, the AIBC seeks to overcome the drawbacks of centralized, non-scalable and high cost traditional cloud computing. AIBC provides a perfect platform for distributed industry solutions (DSOLs) by leveraging the basic blockchain technology and provides system-wide sharing of computing power and storage space.

The AIBC stresses on application support. It provides a flexible technical support infrastructure for distributed services of large business scenarios. Its AI-based fundamental layer Delegated Adaptive Byzantine Fault Tolerance (DABFT) distributed consensus enables technical teams in a variety of industries to focus on their own domain improvement without having to understand the underlying blockchain technology.

The AIBC emphasizes on ecosystem expansion. Our vision is to build a cross-chain, cross-system, cross-industry, cross-application and cross-terminal distributed and trusted ecosystem. Based on an innovative economic model, the AIBC's Delegated Proof of Economic Value (DPoEV) incentive consensus enables connections among diverse computing, data and information entities. Therefore, multi-dimensional business scenarios can be formed with consensus and trust with standalone yet interconnected DSOLs.

The AIBC allows individualized customization based on choices of protocols, modules, and rules. Thus application scenarios in the AIBC ecosystem can be customized according to differentiated requirements of multiple entities on a public chain that provides common bottom-layer services. The customization is from several perspectives, including but not limited to:

1. Technical perspective: The AIBC provides customization based on an entity's technical requirements, such as access mechanism, encryption requirement, consensus (DABFT), storage method, etc.
2. Application perspective: The AIBC provides customization based on an entity's industry standards and guidelines for resource sharing across different domains.
3. Governance perspective: The AIBC provides customization based on the laws, rules, and regulations of the jurisdiction in which an entity resides.

In a nutshell, the mission of the AIBC is to become the value exchange hub with the blockchain technology at its foundation. The value in the AIBC is essentially the knowledge that existed in and accumulated by participating entities. The entities then participate in exchanges of values through resource sharing activities, facilitated by token (unit of economic value) transfers. The benefits of the AIBC are then value creation and exchange across entities.

## 1.2 Blockchain Overview

### 1.2.1 Bitcoin and Ethereum

After the outburst of the 2008 financial crisis, Satoshi Nakamoto publishes a paper titled "Bitcoin: A Point-to-Point Electronic Cash System," symbolizing the birth of cryptocurrencies (Nakamoto, 2008). Vitalik Buterin (Buterin, 2013) improves upon the Bitcoin with a public platform that provides a Turing-complete computing language, the Ethereum, which introduces the concept of smart contracts, allowing anyone to author decentralized applications where they can create their own arbitrary rules for ownership, transaction formats, and state transition functions. The Bitcoin and Ethereum are the first batches of practical blockchain applications that make use of distributed consensus, decentralized ledger, data encryption and economic incentives afforded by the underlying blockchain technology. Essentially, the blockchain technology enables trustless peer-to-peer transactions and decentralized coordination and collaboration among unrelated parties, providing answers to many challenges unsolvable by the traditional centralized institutions, including not but limited to, low efficiency, high cost, and low security.

### 1.2.2. Blockchain Classification

Based on the user's accessibility or the degree of openness, there are three types of blockchains: public chain, coalition chain, and private chain.

A public chain is the most open and anyone can participate in its development and maintenance. Several key benefits of the public chain are: the data is readily accessible by all users; it is easy to deploy applications, and it is completely decentralized without any centralized control. A private chain is the most closed, and its accessibility is limited to the concerned private parties. While a private chain does not completely solve the problem of trust, it improves auditability. A coalition chain is semi-open and requires a registered license to access, thus open to only coalition members. The scale of a coalition can be as large as different institutions and countries.

Table 1.1 compares and contrasts between the three types of blockchains.

Comparison of Three Different Forms of Blockchain

| | Public Chain | Coalition Chain | Private Chain |
|---|---|---|---|
| Participant | anyone | coalition members | personal or internal |
| Consensus mechanism | PoW/PoS/DPoS/PBFT | distributed conformance algorithm | distributed conformance algorithm |
| Bookkeeper | all participants | elected members | user-defined |
| Incentive mechanism | necessary | optional | unnecessary |
| Degree of Centralization | decentralization | multi-centralization | (multi-)centralization |
| Salient Features | self-established credibility | efficiency and cost optimization | transparency and traceability |
| Carrying Capacity | 3-20 deals/s | 1000-1w deals/s | 1000-10w deals/s |
| Typical Scene | virtual digital currency | payment, settlement | audit, issuance |

**Table 1.1 – Comparisons of Blockchains by User Accessibility**

### 1.2.3 Blockchain Technology Architecture

From the perspective of architecture design, the blockchain can be divided into three levels: the protocol layer, the intermediate layer, and the application layer. They are independent but not inseparable.

The protocol layer consists of a complete suite of blockchain protocols; it is similar to a computer operating system. It can be further divided into storage and network layers, where the network nodes are maintained.

The integrity of the protocol layer ensures the high credibility of the system. The protocol layer consists of three components: core technology, core application, and supporting facilities. The core technology component offers the basic protocols and algorithms that the blockchain system depends on, including communication, storage, security and consensus mechanisms. The core application component is built upon the core technology component and provides functions for different application scenarios, such as smart contracts, programmable assets, incentives, etc. The supporting facilities provide resources and tools to the core application component that makes the development process more efficient. Figure 1.1 illustrates the blockchain architecture.

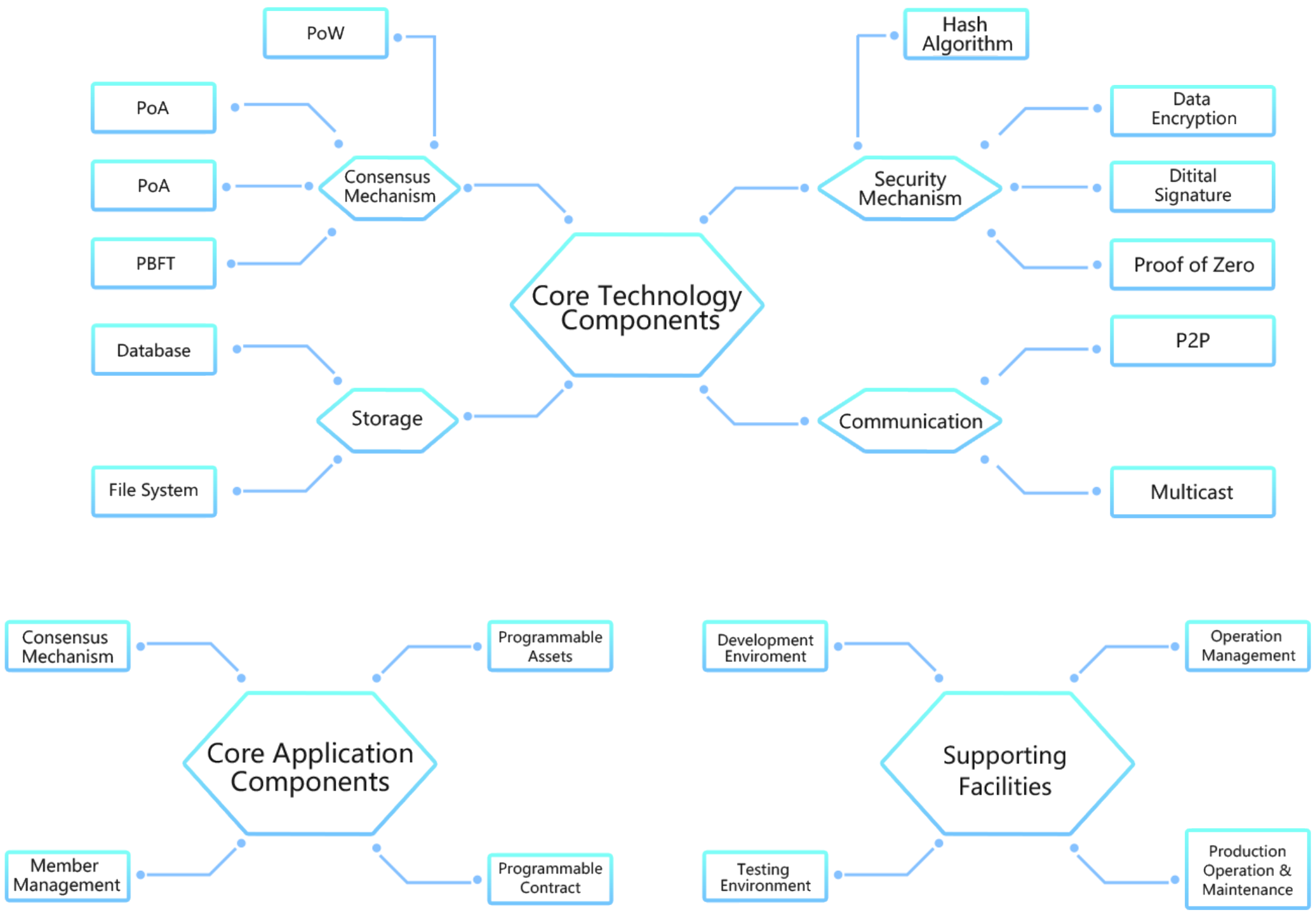

**Figure 1.1 – The Blockchain Architecture**

The intermediate layer is also called the expansion layer, which is further divided into two categories: various trading markets (exchanges) that provides channels for the conversions between cryptocurrencies and legal tenders, and the implementation expansion to certain directions, such as the smart contracts. The application layer is similar to various applications in a computer environment, upon which many applications can be built.

### 1.2.4 Blockchain Ecosystem

Bitcoin, the pioneer of the blockchain's distributed ledger and distributed database revolution, is widely regarded as the "Blockchain 1.0." The "Blockchain 2.0" is represented by Ethereum, which adds a smart contract mechanism to the Bitcoin Foundation. The blockchain is entering its 3.0 era: it is experiencing a proliferation of application scenarios with no apparent scope limitation. It has the potential to become the low-level protocol of the "Internet of Everything." The blockchain applications now cover supply chain finance, transportation, medical and health, culture and media, property right certification, charity and donation management, etc., just to name a few.

From an application scenario perspective, the blockchain technology addresses three core issues, which are explained below:

1. Business applications can benefit from the fact that the data on the chain are of mutual recognition and mutual verification from multiple entities. Thus, data verification costs and security risk for commercial transactions can be significantly reduced, while at the same time the ease of transactions is improved, and the transactions are more deterministic.
2. The blockchain technology is natural for supply chain applications. All participants of a supply chain can help establish and maintain rules and incentive mechanisms, promote collaboration and interoperability, and enhance transparency.
3. The blockchain enables the establishment of distributed databases to solve the trust problem. A blockchain-based database offers trusted and distributed data storage and sharing, is secure and tamper-proof, and incorporates a digital contract platform. Based on these characteristics, the blockchain enables buildup of a social network based, shared data storage, bridging nodes that belong to different entities. This technological approach fundamentally resolves the cross-entity trust issue.

Figure 1.2 illustrates the blockchain ecosystem.

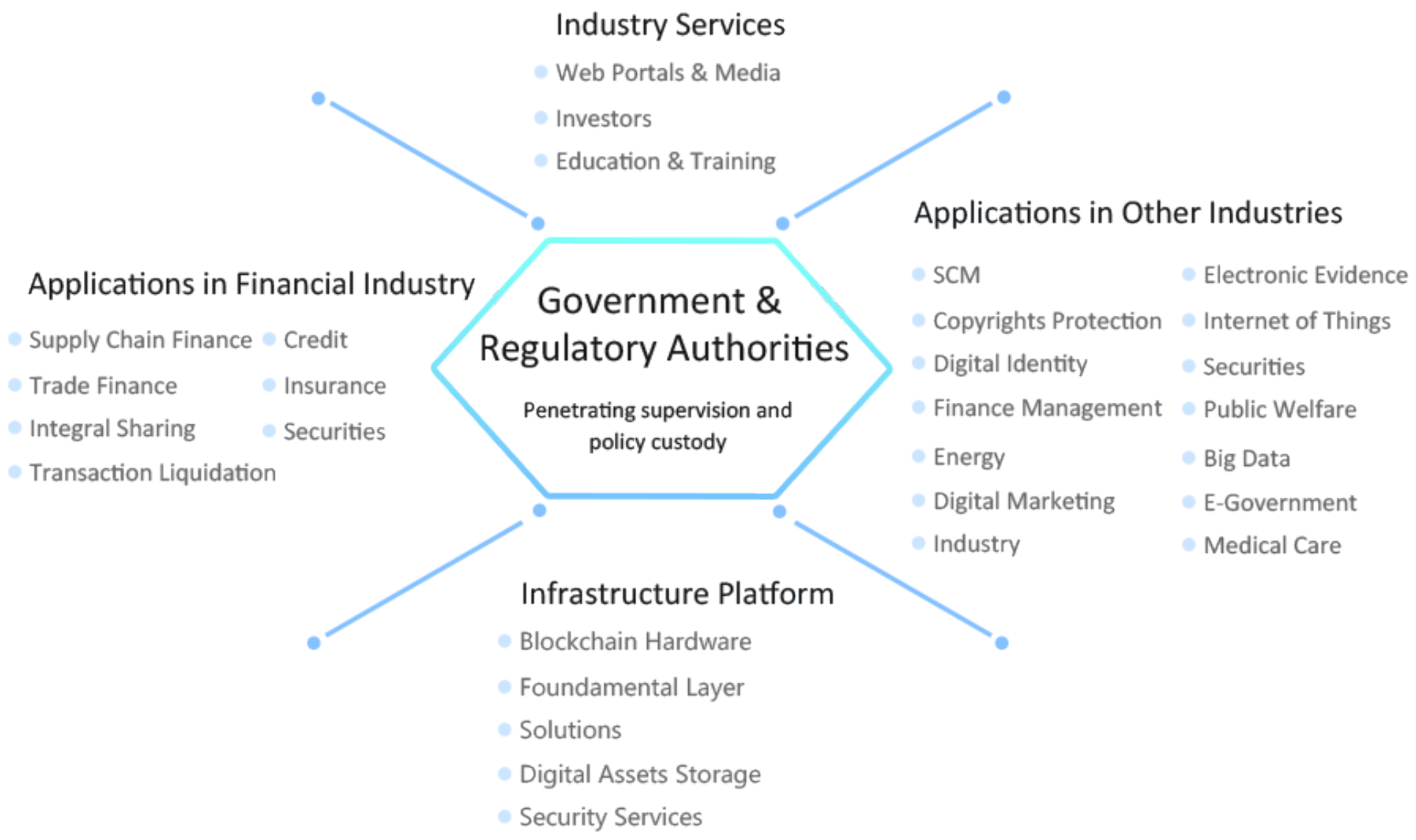

**Figure 1.2 – The Blockchain Ecosystem**

## 1.3 AIBC Overview

### 1.3.1 AIBC Technological Innovations

One of the main innovations of the AIBC is separating the fundamental (blockchain) layer distributed consensus and the application layer incentive mechanism.

Prof. Deng proposed the DABFT (Delegated Adaptive Byzantine Fault Tolerance) as the fundamental layer distributed consensus algorithm. The DABFT improves upon the ADAPT algorithm (Bahsoun, Guerraoui and Shoker, 2015) and uses deep learning (a branch of Artificial Intelligence) techniques to predict and dynamically select the most suitable Byzantine Fault Tolerant (BFT) algorithm for the current application scenario in order to achieve the best balance of performance, robustness and security. The DABFT is currently the most adaptive distributed consensus solution that meets various technical needs among public chains.

At the application level, Prof. Deng proposed an innovative incentive mechanism derived from a model of cooperative economics (macroeconomics, microeconomics, and international trade), the Delegated Proof of Economic Value (DPoEV) consensus. The DPoEV uses the knowledge map algorithm (a branch of Artificial Intelligence) to accurately assess the economic value of digital assets (knowledge).

The AIBC is task-driven with a "blocks track task" dynamic sharding structure. It is designed as a two-dimensional (2D) BlockCloud (as opposed to a 1D blockchain) with super nodes that track the status of tasks through side chains. Once a task is initiated, a set of task validators are then selected according to the "rule of relevancy." A task handler is then chosen among the task validators to handle the task. The task handler and validators manage the task from the beginning to the end with no dynasty change. Thus, effectively, from the task's perspective, the task validators form a shard that is responsible for managing it, with the task handler being its leader.

In essence, a task entails a specific activity initiated by a tasking node in the system, and a blockchain in the AIBC is a side chain consisting of blocks that track the progress of the task (and other tasks) in a dynamically allocated shard in which its task handler is the leader. The 2D dynamic sharding implementation resolves the scalability issue, and at the same time makes it extremely efficient to evaluate the incremental economic value of additional knowledge contributed by each task.

On the ecosystem layer, the AIBC is a "dual-token" platform that marks each decentralized application as a unique entity, yet provides a unified cross-platform value measure. AIBC offers a complete end-to-end distributed industry solution (DSOL). Based on this dual-token platform, AIBC creatively issues (CFTX) tokens for DSOL assets and develop a corresponding cross-chain, permission-based protocol for exchange.

The AIBC combines Artificial Intelligence, big data, cloud computing, and distributed database to provide platforms and algorithms for applications in finance, digital assets, supply chain, education and training, and Internet media, etc.

### 1.3.2 AIBC Business Scenarios

The BlockCloud assets digitization Center provides the first business scenario for AIBC is the Assets Securitization and Tokenization. AIBC provides a decentralized channel for digital assets transactions, effectively transforming these digital assets transactions into social network communications, greatly increasing the flow of digital assets and increasing their value.

The Cofintelligence AI Research Center provides a second business scenario for AIBC: a smart investment and asset management platform. The platform uses AI algorithms (mainly neural networks and deep learning) to analyze the secondary market transactions of different frequencies in order to predict the future trend of financial assets, for the purpose of generating a variety of investment portfolios with different styles. The platform seeks to help institutional and individual investors develop suitable investment strategies.

In addition, the AIBC has the potential to support the following business scenarios:

1. Traceability Application: Through the irreversible, tamperproof and traceable nature of AIBC blockchain, users can effectively trace the authenticity and uniqueness of physical products.
2. Supply Chain Finance: Through the AIBC distributed consensus algorithm, users can effectively match the investment and financing needs in supply chain finance, and manage the credit lines of upstream and downstream enterprises.
3. BlockCloud Service: The architecture of the AIBC itself makes it a good decentralized cloud service platform, which provides distributed ledger recording service and application development environment to enterprise developers.

# 2 AIBC Economic Model Overview

The AIBC ecosystem is essentially a closed economy, of which the operations run on a set of carefully designed economic models. These economic models, at the minimum, must include a macroeconomic model that governs monetary policy (token supply), a trade economic model that enforces fair trade policy, and a microeconomic model that manages supply and demand policy.

## 2.1 Economic Model Overview

### 2.1.1 Macroeconomic Model, Monetary Policy and Digital Money Standard

The most important economic model is the macroeconomic model that provides tools to govern the monetary policy, which principally deals with money (token) supply.

Before the birth of modern central banks, money was essentially in the form of precious metals, particularly gold and silver. Thus, money supply was basically sustained by physical mining of precious metals. Paper money in the modern sense did not come to existence till after the creation of the world's first central bank, Bank of England in 1694. With the creation of the central banks, the modern monetary policy was born. Initially, the goal of monetary policy was to defend the so-called gold standard, which was maintained by their promise to buy or sell gold at a fixed price in terms of the paper money (Abdel-Monem, 2009). The mechanism for the central banks to maintain the gold standard is through setting/resetting the interest rates that they adjust periodically and on special occasions.

However, the gold standard has been blamed for inducing deflationary risks because it limits money supply (Keynes, 1920). The argument gains merit during the great depression of the 1920's and 1930's, as the gold standard might have prolonged the economic depression because it prevented the central banks from expanding the money supply to stimulate the economy (Eichengreen, 1995; AEA, 2001). The "physical" reason behind the gold standard's deflationary pressure on the economy is the scarcity of gold, which limits the ability of monetary policy to supply needed capital during economic downturns (Mayer, 2010). In addition, the unequal geographic distribution of gold deposits makes the gold standard disadvantageous for countries with limited natural resources, which compounds the money supply problem when their economies are in contrarian mode (Goldman, 1981).

An obvious way to combat the gold standard's natural tendency of devaluation risk is to issue paper money that is not backed by the gold standard, the so-called fiat money. A fiat money has no intrinsic value and is used as legal tender only because its issuing authority (a central government or a central bank) backs its value with non-precious metal financial assets, or because parties engaging in exchange agree on its value (Goldberg, 2005). While the fiat money seems to be a good solution for the devaluation problem, central governments have always had a variety of reasons to oversupply money, which causes inflation (Barro and Grilli, 1994). Even worse, as the fiat money has no intrinsic value, it can become practically worthless if the issuing authorities either are not able or refuse to guarantee its value, which induces hyperinflation. Case in point is the Deutsche Mark hyperinflation in the Weimar Republic in 1923 (Federal Reserve, 1943).

Therefore, neither the gold standard nor the fiat currency can effectively create a "perfect" monetary policy that closely matches the money supply with the state of the economy. After the breakdown of the Bretton Woods framework, all economies, developed and developing alike, still, struggle with choices of monetary policy instruments to combat money supply issues de jour. In addition, because of the physical world's "stickiness (of everything)," all money supply instruments (e.g., central bank interest rates, reserve policies, etc.) lag behind the economic reality, making real-time policy adjustment impossible.

Therefore, eradication of deflation and inflation will always be impractical, unless a commodity money with the following properties can be found or created:

1. That it has gold-like intrinsic value but not its physical scarcity.
2. That it can be mined at the exact pace as the economic growth.
3. And that it can be put into and taken out of circulation instantaneously and in sync with economic reality.

Such a commodity does not exist in the physical world. However, things might be different in the digital world, if digital assets can be monetized into digital currencies.

There have been discussions about a "Bitcoin standard." For example, Warren Weber (2015) with the Bank of Canada explores the possibility and scenarios that central banks get back to a commodity money standard, only that this time the commodity is not gold, but Bitcoin. However, just like gold, Bitcoin faces a scarcity challenge in that its quantity is finite, and just like gold it needs to be mined at a pace that may lag far behind economic growth (Nakamoto, 2008). As such, other than that Bitcoin resides in the digital worlds, it does not offer obvious and significant benefits over gold as the anchor for a non-deflationary commodity money standard.

However, such a digital currency can be created, which instantaneously satisfies the requirement that it can be put into and taken out of circulation instantaneously and in sync with economic reality.

The requirements that the digital currency must have gold-like intrinsic value but not its physical scarcity and that it must be mined at the exact pace as the economic growth are not trivial. First of all, there must be an agreement that digital assets are indeed assets with intrinsic value as if they were physical assets. While such an agreement is more of a political and philosophical nature, and therefore beyond the scope of our practicality-oriented interest, it is not a far stretch to regard knowledge as something with intrinsic value, and since all knowledge can be digitized, it thus can form the base of a digital currency with intrinsic value. This is what we call the "knowledge is value" principle.

Based on our "knowledge is value" principle, there is some merit to Warren Buffett's argument that Bitcoin has no intrinsic value, "because [Bitcoin] does not produce anything (Buffett 2018)." Warren Buffett's remarks refer to the facts that during the Bitcoin mining process, nothing of value (e.g., knowledge) is actually produced, and that holding Bitcoin itself does not produce returns the way traditional investment vehicles backed by physical assets do (i.e., through value-added production processes that yield dividends and capital appreciation).

Therefore, again, based on our "knowledge is value" principle, a digital currency that forms the base for a commodity money standard must have intrinsic value in and unto itself; thus not only it is knowledge, it also produces knowledge. This is the fundamental thesis upon which a digital ecosystem that uses a quantitative unit of knowledge as value measurement, thus currency, can be built.

In a digital ecosystem, there is both knowledge in existence and knowledge in production. If the value of knowledge in existence can be directly measured by a quantitative and constant unit, then the unit itself can be regarded as a currency. Furthermore, the value of knowledge in production can also be measured by the constant unit (currency) in an incremental manner, thus expansion of knowledge is in sync with the expansion of currency base. Effectively, the value measurement system is an autonomous monetary policy that automatically synchronizes economic output (knowledge mining) and money supply (currency mining), because the currency is not a stand-alone money, but merely a measurement unit of the value of knowledge. Thus, this digital currency simultaneously satisfies the requirements that it must have gold-like intrinsic value but not its physical scarcity and that it must be mined at the exact pace as the economic growth, as the currency (measurement unit) and the economic growth (knowledge) are now one and the same; they are unified. In the next section, we discuss how to develop the measurement unit.

### 2.1.2 Trade Microeconomic Models and Policy Adjustment

The trade economic model provides tools to enforce fair trade policy among participants in a “globalized” economic environment. In a conventional open and free trade regime with no restrictions, it is quite likely that a few “countries” over-produce (export) and under-consume (import), thus they accumulate vast surpluses with regard to their trading partners. These countries will eventually appropriate all the wealth in the global economy, reducing their trade partners to an extreme level of poverty. Therefore, there must be a fair trade policy, enforced by a collection of bilateral and multilateral trade agreements, which penalizes the parties with unreasonable levels of surplus, and provides incentives to the parties with unreasonable levels of deficit. The penalization can be in the form of tariff levy, and other means to encourage consumption and curb production. The incentives can be tariff credit to encourage production and curb consumption. They are essentially wealth rebalancing devices that a "world trade organization (WTO)" like body would deploy to guarantee that trades should be both free and fair (WTO, 2015).

The microeconomic model provides tools to help manage supply and demand policy in order to set market-driven transaction prices between participants. When there are multiple products simultaneously competing for consumers, the price of a product is set at the point of supply-demand equilibrium. The supply and demand policy discourages initially high-value products to dominate production capability and encourages initially low-value products to be produced. Therefore, consumers can find any product that serves their particular need at reasonable price points.

## 2.2 AIBC Economic Model Implementation Overview

Because of the physical world's "stickiness (of everything)," all monetary policy instruments (e.g., central bank interest rates, reserve policies, etc.), fair trade devices and supply-demand balancing tools lag behind the economic reality. This means these economic models can never dynamically track economic activities and adjust economic policies accordingly on a real-time basis. To make things more complicated, because all economic policies are controlled by centralized authorities (central banks, WTO, etc.), they may not necessarily reflect the best interests of majority participants in economic activities.

The Internet, however, provides a leveling platform that makes real-time economic policy adjustment practical. This is because the digital world can utilize advanced technological tools in order not to suffer from the reality stickiness and policy effect lag that are unavoidable in the physical world, as well as the potential conflict of interest that cannot be systematically eliminated with centralized authorities. The most important tool of them all, in this sense, is the blockchain technology, which provides a perfect platform for a decentralized digital economy capable of real-time economic policy adjustment.

On the upper-layer, the AIBC ecosystem is an implementation of the "knowledge is value" macroeconomic model through an innovative Delegated Proof of Economic Value (DPoEV) incentive consensus algorithm. The DPoEV consensus establishes a digital economy, in which a quantitative unit that measures the value of knowledge, the CFTX token, is used as the media of value storage and transactions. Since the token issuance and the knowledge expansion are unified and therefore always in-sync on a real-time basis, no deflation and inflation exist in the ecosystem by design. Along with the trade and microeconomic models, the AIBC provides a framework of decentralized, consensus-based digital economy with real-time policy adjustment that enables resource sharing.

On the bottom layer, the AIBC implements a Delegated Adaptive Byzantine Fault Tolerance (DABFT) distributed consensus algorithm that enforces the upper-layer DPoEV policies. It combines some of the best features of the existing consensus algorithms and is adaptive, capable of selecting the most suitable consensus for any application scenario. The DABFT is the blockchain foundation upon which the AIBC ecosystem is built.

# 3 AIBC Architecture

## 3.1 AIBC Architecture Overview

The AIBC is an Artificial Intelligence and blockchain technology based decentralized ecosystem that allows resource sharing among participating nodes. The primary resources shared are the computing power and storage space. The goals of the AIBC ecosystem are efficiency, fairness, and legitimacy.

The AIBC consists of four layers: a fundamental layer conducts the essential blockchain functions, a resource layer that provides the shared services, an application layer that initiates a request for resources, and an ecosystem layer that comprises physical/virtual identities that own or operate nodes.

The AIBC implements a two-consensus scheme to enforce upper-layer economic policies and achieve fundamental layer performance and robustness: The Delegated Proof of Economic Value (DPoEV) incentive consensus to create and distribute award on the application and resource layers, and the Delegated Adaptive Byzantine Fault Tolerance (DABFT) distributed consensus for block proposition, validation and ledger recording on the fundamental layer.

The traditional one-dimensional (1D) single-chain ecosystems are not efficient in information retrieval and economic value assessment on the node level. The AIBC is task-driven, and it adopts a concept of "blocks track task." It is designed as a two-dimensional (2D) BlockCloud with side chains that track tasks managed by its super nodes. As a result, it is extremely efficient to evaluate the incremental economic value of additional knowledge contributed by each task.

## 3.2 AIBC Layers

From a technology perspective, the AIBC ecosystem comprises four layers:

1. The fundamental layer (or blockchain layer) that conducts the essential blockchain functions, namely distributed consensus-based block proposition, validation, and ledger recording. The nodes delegated to perform these fundamental blockchain services are the super nodes. The functionality of the super nodes is explained in the next section.

2. The resource layer that provides the essential ecosystem services, namely, computing power and storage space. The AIBC ecosystem is based on the concept that resources are to be shared, and these resources are provided by the computing nodes and storage nodes. While their functions are different, the computing nodes and storage nodes can physically or virtually be collocated or coincide. The functionalities of the computing and storage nodes are explained in the next section.
3. The application layer that requests resources. Each application scenario is initiated by a tasking node. In the AIBC ecosystem, tasking nodes are the ones that have needs for computing power and storage space, thus it is their responsibility to initiate tasks, which in turn drive the generation of economic value. The functionality of the tasking nodes is explained in the next section.
4. The ecosystem layer that comprises physical/virtual entities that own or operate the nodes. For example, a tasking node can be a financial trading firm that needs resources from a number of computing nodes, which can be other trading firms or server farms that provides computing power.

Figure 3.1 illustrates the AIBC layer structure.

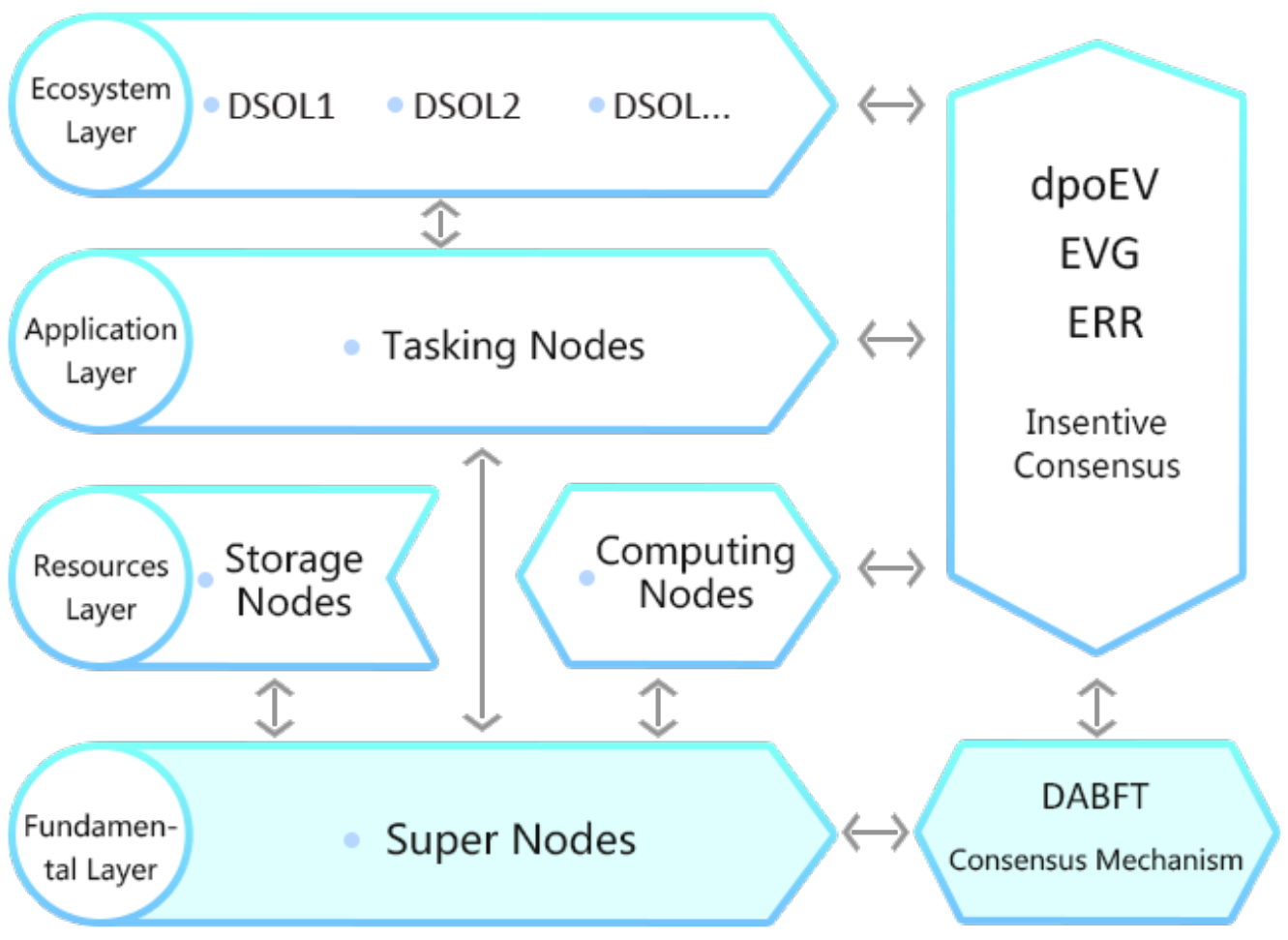

**Figure 3.1 – AIBC Layer Structure**

## 3.3 AIBC Two-Consensus Implementation

The AIBC ecosystem consists of four layers, from top to bottom, they are the ecosystem, application, resource and fundamental (blockchain) layers. These layers have distinguished responsibilities and thus performance/robustness requirements. For example, once a task is initiated, the application and resource layers are primarily concerned with delivering resources and distributing reward. Therefore, these layers need to follow the economic policies strictly and run on a deterministic and robust protocol, but not necessarily a high-performance one (in terms of speed). On the other hand, the fundamental layer is the workhorse providing basic blockchain services such as consensus building, block proposition and validation, transaction tracking, and ledger recording. Therefore it needs to follow an adaptive protocol with high throughput without sacrificing robustness.

As such, the AIBC implements a two-consensus approach: the DPoEV incentive consensus to create and distribute awards on the application and resource layers, and the DABFT distributed consensus responsible for blockchain functions on the fundamental layer. The DPoEV is deterministic and does not necessarily require high-performance as most of the application scenarios do not demand real-time reward distribution. On the other hand, the DABFT has to be real-time and adaptive, as block validation and record bookkeeping need to be done quickly and robustly.

The two-consensus implementation is a distinguishing feature for the AIBC. It enforces upper-layer economic policies and bottom-layer consensus building, a perfect combination for resource-sharing application scenarios. On the other hand, most of the existing and proposed public chains adopt one-consensus schemes, which do not provide flexibility in performance and robustness tradeoff and are vulnerable against risks such as 51%-attacks.

## 3.4 AIBC "Blocks Track Task" Dynamic Sharding Implementation

As the name implies, a blockchain is a chain made of blocks. It is a 1D chain, in which many unrelated transactions are packed in the same block that is then linked up with other unrelated blocks. Therefore, in a single-chain ecosystem such as Bitcoin and Ethereum, if there is a need to track all transactions originated from a node, a rather time-consuming chain-wide search has to be conducted. While this may not appear to be fatal if the only purpose is mining blocks in order to earn ledger recording right and token rewards (like in Bitcoin), it is severely inefficient for any real application (DSOL) scenarios that need to access the information in, and assess the economic value of any particular node. In addition, another main challenge the single-chain ecosystems face is scalability, which is key to performance improvement.

To resolve the inefficiency problem that the traditional 1D single-chain ecosystems face, the AIBC is designed as a 2D BlockCloud, in which all chains are side chains consisting of only related blocks (blocks managed by the same super node). The AIBC is task-driven, and it adopts a concept of "blocks track task." When a task is initiated by a tasking node, the side chain attached to the super node elected to handle the task is extended. The side chain grows with each additional block tracking the status of the task until it is completed and closed permanently. Thus, the side chain contains all static and dynamic information of the task, as well as additional knowledge brought about by the task. As the super node continues to manage tasks, the side chain grows. As such, any business application scenarios that need to access the information in, and assess the incremental economic value of any super node can do so without any inefficient system-wide search, as the information is readily available in the side chains that track tasks handled by that node. Moreover, because of the AIBC's 2D construct, information access and economic value assessment can be done in parallel on many nodes, which further improves the efficiency of the ecosystem.

The AIBC ecosystem also resolves the scalability problem with a dynamic sharding feature by design. In the AIBC 2D BlockCloud, once a task is initiated, a set of task validators are then selected according to the "rule of relevancy," among which a task handler is then chosen. Thus from a task's perspective, the task validators form a shard with the task handler being its leader. Because of the "rule of relevancy," it is highly likely that each new task is assigned a different set of task validators from the previous task.

Therefore, once a task is completed, its shard dissolves automatically, and no periodically "re-sharding" is necessary. Such dynamic sharding feature makes the AIBC easy to scale with yet further improved efficiency.

Figure 3.2 illustrates the AIBC side chain implementation.

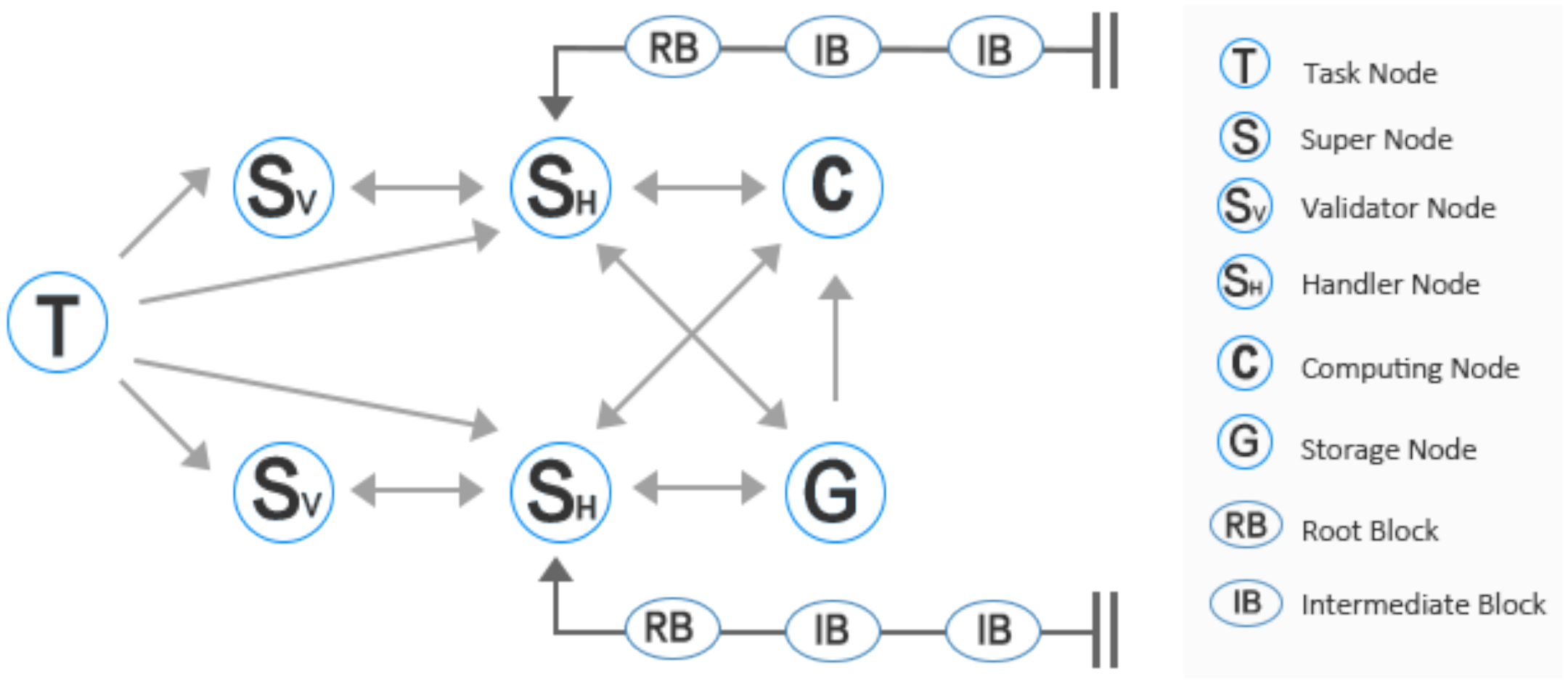

**Figure 3.2** – **AIBC Side Chain Implementation**

Figure 3.3 and Figure 3.4 illustrate the architectures of traditional sharding and the AIBC dynamic sharding, respectively.

The AIBC also maintains a 1D "main chain" at each super node. The 1D AIBC "main chain" is essentially a flattened representation of the 2D cloud, with the blocks of side chains from all super nodes intertwined. A Merkle tree structure of the 1D blockchain makes it topologically identical to the 2D BlockCloud.

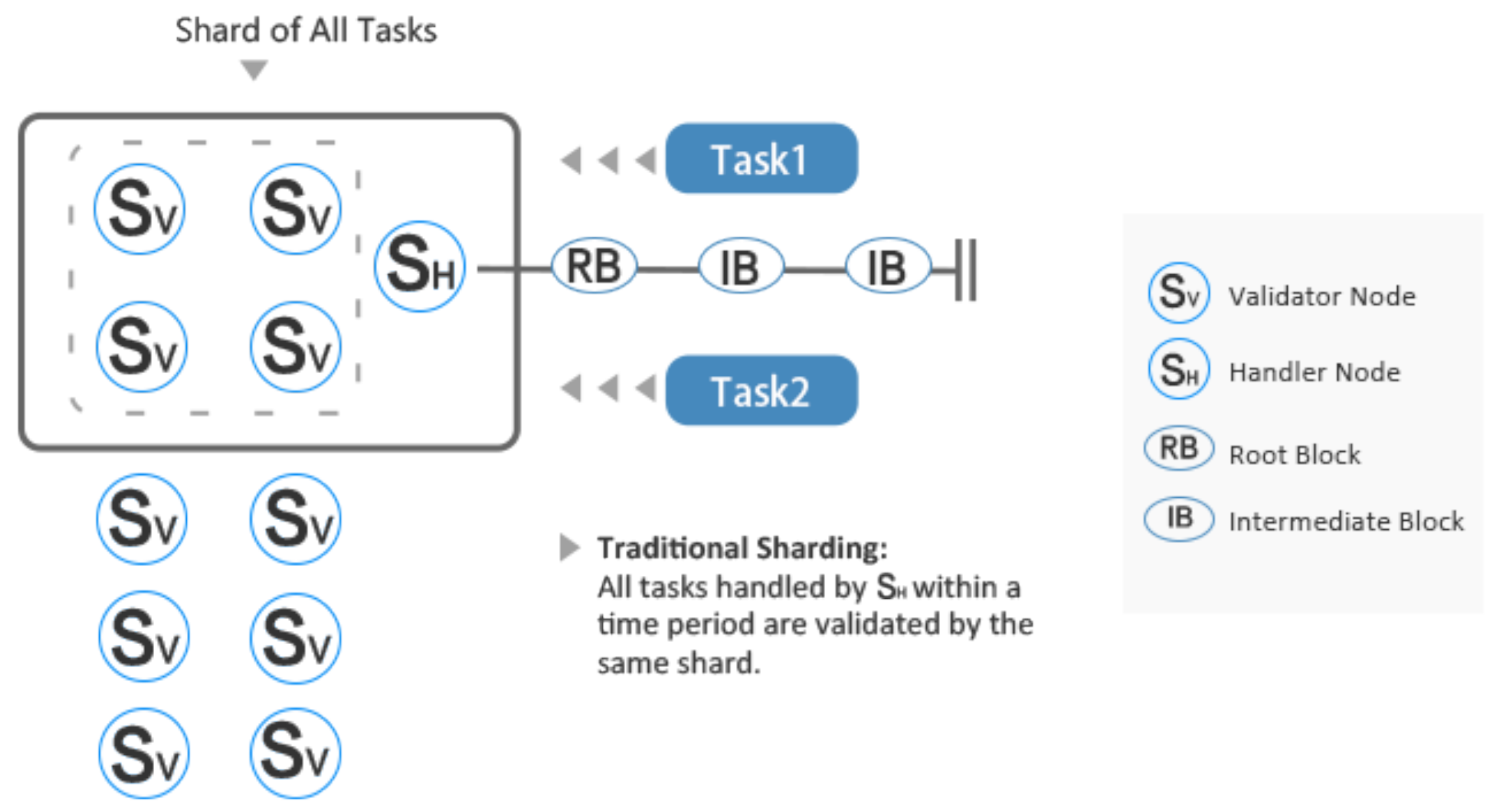

**Figure 3.3 – Architecture of Traditional Sharding**

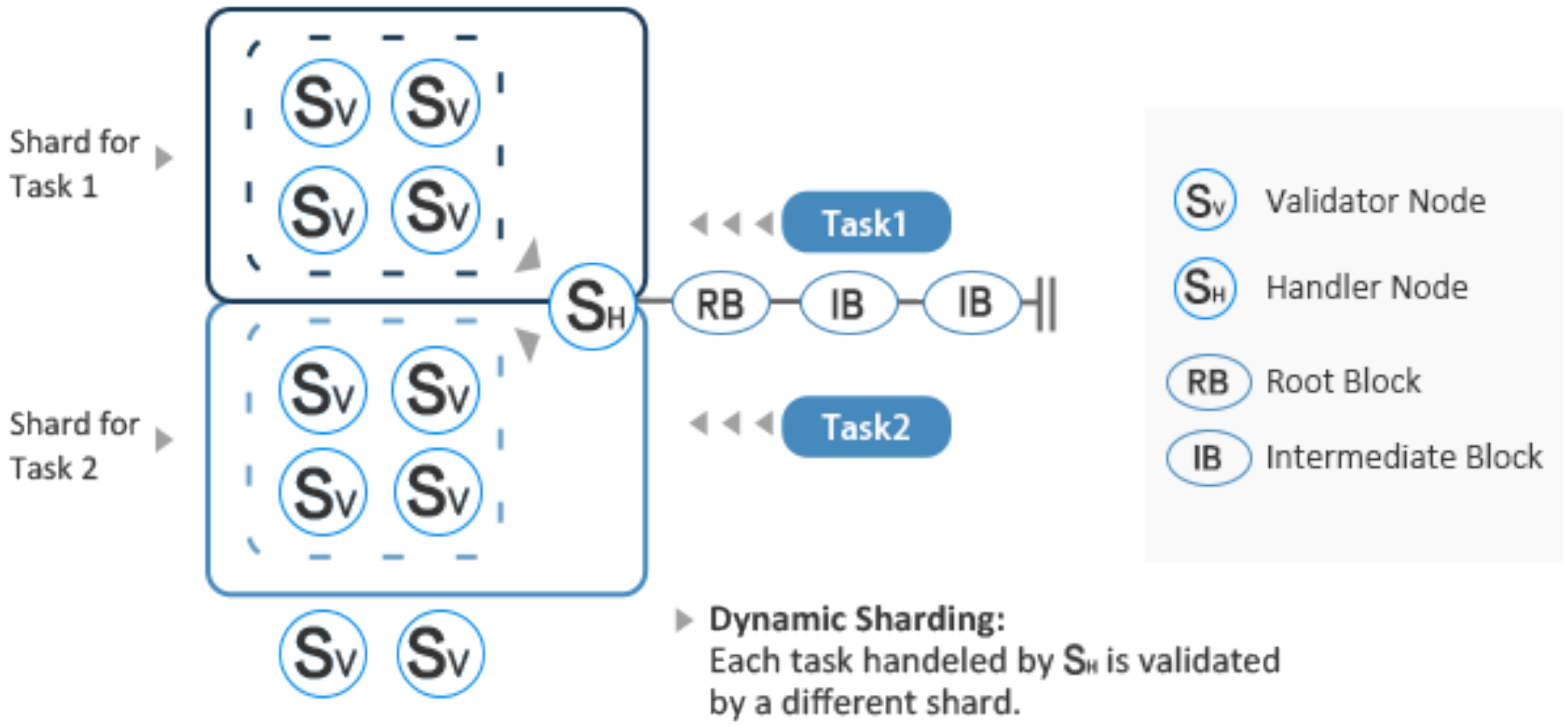

**Figure 3.4 – Architecture of AIBC Dynamic Sharding**

## 3.5 AIBC Dual-Token DSOL Platform

On the ecosystem layer, the AIBC is a "dual-token" platform that marks each DSOL as a unique and identifiable entity, yet provides a unified cross-platform value measure. In the entire ecosystem, In addition to the system-wide unified measure of value and transaction medium CFTX, each DSOL is issued a number (e.g., 1,000) of its own distinguishable tokens, the DSOLxxxx. The dual-token approach allows the CFTX be used for the entire AIBC ecosystem, while enables the transfer of DSOL ownership on a whole sale level through auctions of DSOLxxxx tokens.

## 3.6 AIBC Asset Anchoring

Based on the dual-token platform, AIBC creatively issue the token with asset anchoring value (CFTX). Different from the existing "Valid Tokens", all the tokens in AIBC form a one-to-one binding relationship with assets through smart contracts, and all asset packages have legal credentials as the basis. AIBC achieves this by allowing DSOL owners in the ecosystem to create and exchange tokens within their DSOL environment. The anchor token in DSOL is created when the DSOL asset is mortgaged or injected, and the currency denominated assets are 1:1 exchanged into the system to anchor the certificate. When the value of the asset changes, the corresponding change occurs through the smart contract anchor. The number of anchor token created and issued must never exceed the underlying asset value. The AIBC solution provides both a secure offline approval mechanism and a flexible online approval mechanism, all controlled by smart contracts.

## 3.7 AIBS Permission-based Cross-chain Exchange Protocol

In the AIBC ecosystem, multiple DSOLs based on the AIBC underlying public chain can be understood as independent economies. The communication between the various DSOLs and their communication with the underlying AIBC public chain (and future communication between AIBC and other public chains), especially sensitive information, must be addressed. For now, there is no clear and uniform exchange standard for cross-chain exchange. Moreover, various cross-chain methods that existed are for the token of no-collection attributes, and there is no cross-chain exchange protocol for the collection property. While AIBC adopted a standard protocol for permission-based cross-chain exchange to solve these problems.

# 4 Delegated Proof of Economic Value (DPoEV)

## 4.1 DPoEV Overview

Inside the AIBC ecosystem, all activities create (or destroy) economic value. Therefore, there is a need for a logical and universal way to assess the economic value of an activity, measured by the community's value storage and transaction medium, the CFTX token. The DPoEV incentive consensus algorithm is to create and distribute award to participating nodes in the AIBC ecosystem. The DPoEV, in turn, is established upon an innovative Economic Value Graph (EVG) approach, which is derived from the knowledge graph algorithm (a branch of Artificial Intelligence and deep learning). The EVG is designed to measure the economic value ("wealth") of the ecosystem in a dynamic way. The EVG will be explained in the next sub-section.

The implementation of DPoEV is as follow:

1. At the genesis of the AIBC, the EVG mechanism accurately assesses the economic value, or initial wealth ("base wealth") of the knowledge in the entire ecosystem (all participating nodes: super nodes, tasking nodes, computing nodes and storage nodes to be explained in the next few sections), and comes up with a system-wide wealth map. The DPoEV then issues an initial supply of CFTX tokens according to the assessment.
2. Afterward, the EVG updates the wealth map of the entire ecosystem on a real-time basis, with detailed wealth information of each and every node in the ecosystem. In the AIBC ecosystem, wealth generation is driven by tasks. The EVG assesses the incremental wealth brought about by a task, and the DPoEV issues fresh tokens accordingly. This enables the ecosystem to dynamically adjust the money (token) supply to prevent any macroeconomic level deflation and inflation in a very precise manner. Essentially, the DPoEV supervises monetary policy in a decentralized ecosystem.
3. The DPoEV monitors the real-time transactions among participating nodes of a task and manages the token award mechanism. After an amount of tokens is created for a task, the DPoEV distributes tokens to nodes that participate in the task. The number of tokens awarded to each node, as well as transaction costs (gas) attributed to the node, depending on that node's contribution to the task.

4. In an open and free trade economic system with no restrictions, it is quite likely that a few resource nodes will accumulate a tremendous level of production capability (computing power and storage space) and experience (task relevancy), who may then be given a majority of tasks/assignments due to a "rule of relevancy" ranking scheme. This would accelerate wealth generation for these dominating nodes in a speed that is unfair to other resource nodes. This is where a "rule of wealth" scheme comes in as a counter-balance, as the DPoEV can elect to grant assignments to nodes with lowest levels of wealth. If, however, there are simply not enough low wealth level resource nodes, which renders the "rule of wealth" ineffective, a "rule of fairness" scheme then comes to play. The "rule of fairness" imposes tariff levies on the dominating nodes, which are then distributed to resource nodes with a low probability of winning assignments. Thus, the DPoEV also functions as a "world trade organization" that enforces fair trade in a decentralized ecosystem.
5. When there are multiple tasks on the ecosystem simultaneously competing for limited resources, the DPoEV decides on a real-time basis whether and how to adjust the value of each task, based on factors such as that how many similar tasks have been initiated and completed in the past and the historical values of these tasks. This prevents initially high-value tasks dominating the limited resources and encourages initially low-value tasks to be proposed. Thus, on the microeconomic level, the DPoEV dynamically balances the supply and demand of tasks. If, in rare cases, the aggregated outcome of task value adjustments is in conflict with the macroeconomic level goal (no inflation or deflation), a value-added tax (VAT) liability (in case of inflation) or a VAT credit (in case of deflation) can be posted on a separate ledger, of which the amount can be used to adjust the next round of macroeconomic level fresh token issuance. Thus the DPoEV provides a central-bank-like open market operations service in a decentralized ecosystem.

6. Finally, the DPoEV conducts periodical true-up as an extra layer of defense for housekeeping purposes. One of the key activities during true-up is for the DPoEV to "burn" surplus tokens that have been created, however, have not been awarded to participating nodes because of economic policy constraints. This is somewhat equivalent to central banks' action of currency withdrawal, which is a macroeconomic tool to destroy currency with low circulation efficiency.
7. The DPoEV is essentially conducted by the super nodes to ensure performance and efficiency in the ecosystem, this is where the "D (Delegated)" in DPoEV comes from.

## 4.2 Economic Value Graph (EVG) Overview

Up to this point, we still have not answered the question of how the value of knowledge is actually measured. The pursuit of a public blockchain is to create an ecosystem that supports a variety of application scenarios, and one of the challenges is to define a universal measurement of economic value.

We propose an innovative Economic Value Graph (EVG) mechanism to dynamically measure the economic value ("wealth") of knowledge in the AIBC ecosystem. The EVG is derived from the knowledge graph algorithm, which is very relevant in the context of the AIBC.

### 4.2.1 Knowledge Graph Overview

A knowledge graph (or knowledge map) consists of a series of graphs that illustrate the relationship between the subject's knowledge structure and its development process. The knowledge graph constructs complex interconnections in a subject's knowledge domain through data mining, information processing, knowledge production and measurement in order to reveal the dynamic nature of knowledge development and integrate multidisciplinary theories (Watthananona and Mingkhwanb, 2011).

A knowledge graph consists of interconnected entities and their attributes; in other words, it is made of pieces of knowledge, each represented as an SPO (Subject-Predicate-Object) triad. In knowledge graph terminology, this ternary relationship is known as Resource Description Framework (RDF). The process of constructing a knowledge graph is called knowledge mapping.

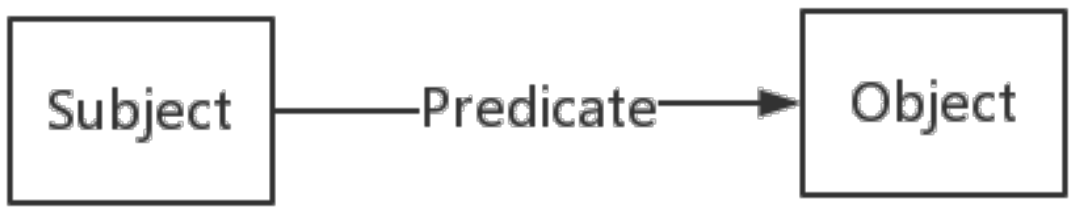

**Figure 4.1 – Knowledge Graph Subject-Predicate-Object Diagram**

The knowledge graph algorithm is consistent with the EVG. There are two steps in knowledge mapping for an ecosystem: realization of initial knowledge, and dynamic valuation of additional knowledge.

For an ecosystem, at the realization of initial knowledge stage, the knowledge graph algorithm assesses *i*th node's initial economic value of knowledge, which is a combination of explicit and implicit economic values of all relevant knowledge pieces at and connected to that node. The total economic value of the entire ecosystem is thus the sum of all node level economic values.

$$v0_i = \prod_{j=1}^{M} P(v0_{i,j}|v0_{i,j-1})\, v0_{i,j} \tag{4.1a}$$

$$V0 = \sum_{i=1}^{N} v0_i \ , \ i = 1, \dots, N, \ j = 1, \dots, M \tag{4.1b}$$

Where $v0_{i,j}$ is the initial economic value of the *j*th knowledge piece, and $P(v_{i,j}|v_{i,j-1})$ is the probability of $v0_{i,j}$ given all knowledge pieces prior to the *j*th, at the *i*th node, and $\Pi$ is a Cartesian product.

Once the initial economic value of the ecosystem is realized, in a task-driven ecosystem, as the tasks start to accumulate, a collection of knowledge graphs of the tasks is then created to assess the incremental economic value of the new knowledge. Finally, the knowledge graph of the entire ecosystem is updated. This dynamic valuation of additional knowledge requires automatic extraction of relationships between the tasks and participating nodes, as well as relationship reasoning and knowledge representation realization. The total economic value of the entire ecosystem is thus the sum of all node level updated economic values.

$$t1_i = \prod_{k=1}^{K} P(t1_{i,k}|t1_{i,k-1})t1_{i,k} - \prod_{k=1}^{K} C(t1_{i,k}|t1_{i,k-1})t1_{i,k} \quad (4.2a)$$

$$T1 = \sum_{i=1}^{N} t1_i \quad (4.2b)$$

$$V1 = V0 + T1 = \sum_{i=1}^{N}( v0_i + t1_i), i = 1, \dots, N, \ k = 1, \dots, K \quad (4.2c)$$

Where $t0_{i,k}$ is the incremental economic value of the *k*th knowledge piece of the task, $P(t1_{i,k}|t1_{i,k-1})$ is the probability of $t1_{i,k}$ given all knowledge pieces prior to the *k*th, $C(t1_{i,k}|t1_{i,k-1})$ is the covariance of $t1_{i,k}$ given all knowledge pieces prior to the *k*th, at the *i*th node, and $\Pi$ is Cartesian product.

### 4.2.2 EVG Implementation

The essence of EVG is "knowledge is value," and it accesses the entire ecosystem's economic value dynamically.

At the genesis of the AIBC ecosystem, there are no side blockchains, as no task has been initiated yet, and the EVG mechanism just simply depicts a knowledge graph of each and every node (super, tasking, computing and storage node) in the blockchain. The EVG then aggregates the knowledge graph of all nodes and establishes a global knowledge graph. At this juncture, the EVG has already assessed the original knowledge depository of the entire ecosystem. Furthermore, in order to quantify this original wealth, the EVG equates it to an initial supply of CFTX tokens, issued by the DPoEV consensus. This process establishes a constant measurement unit of economic value (token) for the future growth of the ecosystem. The EVG then creates a credit table, which contains all nodes in the ecosystem, and their initial economic values. When a new node joins the ecosystem, the EVG appends a new entry to the credit table for it, with its respective initial economic value. The credit table resides in all super nodes, and its creation and update need to be validated and synchronized by all super nodes by the fundamental layer DABFT distributed consensus algorithm. The DABFT will be discussed in the next section.

The wealth generation is driven by tasks, and the super nodes are the ones that are responsible for handling them. As the tasks continue to be initiated, side chains continue to grow and accumulate from the super nodes. These side chains are the containers of the incremental knowledge, and the EVG measures the economic value of this incremental knowledge with the measurement unit (token). Upon the acceptance of every task, the DPoEV consensus issues a fresh supply of CFTX tokens proportional to the newly created economic value to ensure that the money supply is in sync with the economic growth in order to avoid macroeconomic level inflation or deflation.

Each task is tracked by a distinguished task blockchain, which is a side chain with the root block connected to its handling super node. Each block in the task blockchain tracks the status of the task. The root block contains information including the initial estimation of the economic value of the task. Each subsequent block provides updated information on contributions from the task validation, handling, and resource nodes. When the task blockchain reaches its finality, the EVG has a precise measure of economic value generated by this task. Furthermore, the blocks contain detailed information on contributions from participating nodes, and transactions. Thus, the EVG can accurately determine the size of the reward (amount of tokens) issued to each participating node. The DPoEV then credits a respective amount of tokens to each participating node, which is recorded in the credit table validated by the DABFT consensus.

The EVG enables the DPoEV to manage the economic policy of the ecosystem on a real-time basis through the credit table. The DPoEV can dynamically determine the purchase price of a task, which covers the overall cost paid to the super and resource nodes. It can also set the transaction cost for each assignment. The overall effect is that all macroeconomic, microeconomic and trade policies are closely monitored and enforced.

| Economic Value (CFTX Token) | Total Economic Value | Initial Economic Value | Incremental Economic Value - Task 1 | Incremental Economic Value - Task k | Incremental Economic Value - Task K |
|---|---|---|---|---|---|
| Super Node 1 | 1,250,000 | 1,000,000 | 1000 | 750 | 500 |
| … | … | … | … | … | … |
| Super Node $N_S$ | 2,100,000 | 2,000,000 | 2000 | 1500 | 1000 |
| Tasking Node 1 | 75,000 | 50,000 | 50 | 30 | 25 |
| … | … | … | … | … | … |
| Tasking Node $N_T$ | 125,000 | 75,000 | 75 | 60 | 50 |
| Computing Node 1 | 200,000 | 100,000 | 100 | 90 | 75 |
| … | … | … | … | … | … |
| Computing Node $N_C$ | 300,000 | 250,000 | 250 | 175 | 100 |
| Storage Node 1 | 200,000 | 150,000 | 150 | 80 | 25 |
| … | … | … | … | … | … |
| Storage Node $N_{ST}$ | 350,000 | 300,000 | 300 | 175 | 75 |

**Table 4.1 – EVG Node Credit Table**

## 4.3 Economic Relevancy Ranking (ERR)

While the EVG measures the economic value of knowledge created by task, it does not assess the validation, handling, computing, and storage capabilities of participating nodes, as these capabilities are not necessarily based on knowledge. This can be fatal because the DPoEV assigns tasks to super nodes and resource nodes first and foremost with a "rule of relevancy" ranking scheme. This issue is resolved by the Economic Relevancy Ranking (ERR) mechanism.

The ERR ranks tasks as well as the super node and resource nodes (collectively known as "service nodes"). Based on the ERR rankings, the DPoEV provides a matchmaking service that pairs tasks and service nodes.

The ERR assesses each newly created task by the following factors:

1. Time criticalness: How much time a task takes before a task timer expires.
2. Computing intensity: How much computing power is required to complete the task and associated assignments.

3. The Frequency of transactions: Higher transaction frequency improves liquidity, which further increases transaction frequency. Higher transaction frequency allows a faster growth of wealth, however, brings higher demand to the network and database framework.
4. The Scale of transactions: Larger transaction scale improves liquidity, which further increases the transaction scale. Larger transaction scale allows a faster growth of wealth, however, brings higher demand to the network and database framework.
5. Required propagation: Stronger propagation in terms of bandwidth means improved liquidity, which improves transaction frequency and scale. Stronger propagation allows a faster growth of wealth, however, brings higher demand to the network and database framework.
6. Optional data requirement: What and how much data is required to complete the task and associated assignments, and more importantly, where the data is stored.

The ERR ranking score of a task is thus given as

$$TR_{ERR} = \sum_{i=1}^{N} \frac{w_i TR_i}{n_i}, \sum_{i=1}^{N} w_i = 1 \quad (4.3)$$

Where $TR_i$ is ranking score of the *i*th factor, $w_i$ is that factor's weight, and $n_i$ is the factor's normalization coefficient.

As tasks start to accumulate, they are ranked by the above criteria. The ERR then creates a task ranking table, which contains the addresses of all tasks (root blocks of side chains) and their ranking scores. When a new task is initiated, the ERR appends a new entry to the task ranking table for it, with its respective ranking score. The task ranking table resides in all super nodes, and its creation and update need to be validated and synchronized by all super nodes by the DABFT consensus.

| Economic Relevancy Ranking (ERR Score) | | Task 1 | … | … | Task i | … | … | Task N |
|---|---|---|---|---|---|---|---|---|
| | | 0.39 | | | 0.59 | | | 0.46 |
| Time Criticalness | Score | 100 | … | … | 50 | … | … | 75 |
| | Weight | 0.15 | | | 0.05 | | | 0.25 |
| | Normalization Coefficient | 100 | | | 100 | | | 100 |
| Computing Intensity | Score | 25 | … | … | 75 | … | … | 100 |
| | Weight | 0.15 | | | 0.25 | | | 0.15 |
| | Normalization Coefficient | 100 | | | 100 | | | 100 |
| Frequency of Transactions | Score | 5,000 | … | … | 250,000 | … | … | 100,000 |
| | Weight | 0.25 | | | 0.25 | | | 0.15 |
| | Normalization Coefficient | 1,000,000 | | | 1,000,000 | | | 1,000,000 |
| Scale of Transactions | Score | 5 | … | … | 8 | … | … | 3 |
| | Weight | 0.25 | | | 0.25 | | | 0.15 |
| | Normalization Coefficient | 10 | | | 10 | | | 10 |
| Required Propagation | Score | 350 | … | … | 500 | … | … | 150 |
| | Weight | 0.15 | | | 0.15 | | | 0.15 |
| | Normalization Coefficient | 1,000 | | | 1,000 | | | 1,000 |
| Data Requirement | Score | 50 | … | … | 75 | … | … | 25 |
| | Weight | 0.05 | | | 0.05 | | | 0.15 |
| | Normalization Coefficient | 100 | | | 100 | | | 100 |

**Table 4.2 – ERR Task Ranking Score Table**

In parallel, the ERR assesses the capabilities of the service nodes based on the same criteria. It then creates a service node ranking table, which contains the addresses of all service nodes and their ranking scores. When a new service node joins, the ERR appends a new entry to the service node ranking table for it, with its respective ranking score. The service node ranking table resides in all super nodes, and its creation and update need to be validated and synchronized by all super nodes with the DABFT consensus.

The ERR algorithm has three major properties:

1. Consistency. A ranking score, once recorded, cannot be altered through paying more cost by the tasking node (for task ranking) or the service node (for service node ranking). However, the ranking score does change as both the tasking and service nodes do evolve. Adjustment to the ranking score can only be conducted by DPoEV through the DABFT consensus.

2. Computability. The ERR ranking scores need to be retrieved by the DPoEV instantly, thus the ERR algorithm requires low computational complexity.
3. Deterministicness. The ERR algorithm should produce identical results on all nodes for the same service node.

The ERR ranking score of a service node is thus given as:

$$SNR_{ERR} = \sum_{j=1}^{M} \frac{w_j SNR_j}{n_j}, \sum_{j=1}^{M} w_j = 1 \tag{4.4}$$

Where $SNR_j$ is the ranking score of the $j$th property, $w_j$ is that property's weight, and $n_j$ is that property's normalization coefficient.

Based on the ERR ranking scores of tasks and service nodes, the DPoEV provides a matchmaking service that pairs tasks with service nodes with the closet ranking scores. Thus the "rule of relevancy" in service node selection is observed, and service nodes with the highest rankings cannot dominate task handling and assignment. Rather, they have to be "relevant" to the tasks for which they compete. In addition, the "rule of wealth" and "rule of fairness" are used to enforce economic principles.

| Economic Relevancy Ranking (ERR Score) | | Super Node 1 | … … | Computing Node 1 | … … | Service Node 1 |
|---|---|---|---|---|---|---|
| | | 0.50 | | 0.35 | | 0.70 |
| Consistency | Score | 50 | … … | 25 | … … | 35 |
| | Weight | 0.50 | | 0.70 | | 0.25 |
| | Normalization Coefficient | 100 | | 100 | | 100 |
| Computability | Score | 50 | … … | 60 | … … | 75 |
| | Weight | 0.30 | | 0.20 | | 0.30 |
| | Normalization Coefficient | 100 | | 100 | | 100 |
| Deterministicness | Score | 50 | … … | 50 | … … | 85 |
| | Weight | 0.20 | | 0.10 | | 0.45 |
| | Normalization Coefficient | 100 | | 100 | | 100 |

**Table 4.3 – ERR Service Node Ranking Score Table**

The service node selected (out of N service nodes) given a task *j* is follows the following equation:

$$SN_T = min[\cup_i^N (SN_{ARR,i} - TR_{ERR,j})] \tag{4.5}$$

Where $SN_{ERR,i}$ is the ERR ranking score of the *i*th service node, and $TR_{ERR,j}$ is the ERR ranking score of the *j*th task.

It is important to notice that, unlike the EVG, the ERR does not measure the economic value of tasks and service nodes. Rather, it ranks them based on their requirements and capabilities, which are not the bearers of economic value, but its producers. As such, the ERR has no role in money supply policy in the DPoEV framework.

## 4.4 DPoEV Advantages

The DPoEV incentive consensus algorithm creates and distributes award to participating nodes in the AIBC ecosystem in the form of CFTX tokens. It eliminates the possibility of macroeconomic level inflation and deflation, enforces free and fair trade, and balances microeconomic level supply and demand.

With the EVG and ERR, by design, the DPoEV enforces the economic policies and the "rules of relevancy, wealth and fairness." It thus guarantees that no tasking nodes can dominate task initiation, no super nodes can dominate task handling, and no resource nodes can dominate task assignment.

A key benefit of the DPoEV is that it effectively eliminates the possibility of 51% attack based on the number of efforts (like Proof-of-Work in Bitcoin), or wealth accumulation (like Proof-of-Stake in Ethereum). As a matter of fact, it has the potential to eliminate 51% attack of anything.

# 5 Delegated Adaptive Byzantine Fault Tolerance (DABFT)

While the DPoEV algorithm provides the application layer incentive consensus, it needs to work with a high-performance fundamental layer distributed consensus protocol that actually provides blockchain services. This bottom layer consensus is the "real" blockchain enabler.

Therefore, unlike most of the existing public chains, the AIBC establishes a two-consensus approach: on the application layer, the DPoEV consensus is responsible for economic policy enforcement, and on the fundamental layer, a Delegated Adaptive Byzantine Fault Tolerance (DABFT) distributed consensus algorithm is responsible for managing each and every transaction in terms of block generation, validation, and ledger recording. While the DPoEV does not need to be real-time as most of the application scenarios do not demand real-time reward distribution, the DABFT has to be real-time, as block validation and ledger recording need to be done quickly and robustly. The goal of DABFT is to achieve up to hundreds of thousands of TPS (Transactions per Second).

## 5.1 DABFT Design Goals

The DABFT implements the upper-layer DPoEV economic policies on the fundamental layer and provides the blockchain services of block generation, validation, and ledger recording. It focuses on the AIBC's goals of efficiency, fairness, and legitimacy. Unlike the dominant consensus algorithms (e.g., PoW) that waste a vast amount of energy just for the purpose of winning ledger recording privilege, the DABFT utilizes resources only for meaningful and productive endeavors that produce economic value.

## 5.2 Major Consensus Algorithms and DABFT

The DABFT is proposed as all the existing blockchain consensus algorithms do not sufficiently meet the AIBC goals and hence, do not comply with the DPoEV economic models.

### 5.2.1 PoW (Proof of Work) Workload Proof Consensus

The PoW consensus behind Bitcoin plays the zero-sum game of SHA256 hash for the miners to win ledger recording privilege. With the increased level of difficulty on block mining, the PoW wastes a tremendous amount of computing power (and electricity) with a great reduction of throughput. Even worse, the higher number of miners, the higher level of difficulty of mining, and lower level of probability for a miner to win ledger recording privilege induce a yet higher level of energy waste and longer latency. This is the key reason why Ethereum has long considered the use of the PoS (Proof-of-Stake) algorithm Casper instead of the PoW. Therefore, from the perspective of mining speed and cost, the PoW is not conducive to long-term and rapid development of blockchain based ecosystems, and is not consistent with the AIBC goal of efficiency (high-performance) and the DPoEV requirement of "rule of fairness."

### 5.2.2 PoS (Proof of Stake) Equity Proof Consensus and DPoS

The PoS consensus measures the amount and age of wealth in the ecosystem in order to grant ledger recording privilege (Buterin, 2013). PeerCoin (King and Nadal, 2012), NXT (NXT, 2015), as well as the Ethereum's Casper implementation (Buterin, 2014), adopt the PoS. Although the PoS consumes a much lower level of energy than the PoW, it amplifies the impact of accumulated wealth, as such, in a PoS ecosystem, participants with a higher level of wealth can easily monopolize ledger recording. In addition, block confirmations are probabilistic, not deterministic, thus in theory, a PoS ecosystem may have exposure to other attacks. Therefore, from the perspective of miner composition, the PoS is not conducive to the interests of participants in the ecosystem, and is not consistent with the AIBC goal of fairness and the DPoEV requirements of being deterministic, as well as "rule of wealth" and "rule of fairness."

The DPoS is derived from the PoS, and is being used by EOS (EOS, 2018). The main difference is that, in the DPoS regime, all asset holders elect a number of representatives, and delegate consensus building to them. The regulatory compliance, performance, resource consumption, and fault tolerance of the DPoS are similar to that of the PoS. The key advantage of the DPoS is that it significantly reduces the number of nodes for block verification and ledger recording, thus is capable of achieving consensus in seconds.

### 5.2.3 PoI (Proof of Importance) Importance Proof Consensus

The PoI introduces the concept of account importance, which is used as a measure to allocate ledger recording privilege (NEM, 2018). The PoI partly resolves the wealth monopolization dilemma of the PoS. However, it exposes to a nothing-at-stake scenario, which makes cheating rather low cost. Therefore, the PoI deviates from the AIBC goal of legitimacy and the DPoEV requirement of "rule of relevancy."

### 5.2.4 PoD (Proof of Devotion) Contribution Proof Consensus

The PoD introduces the concept of contribution and awards ledger recording privilege according to contributions of accounts (NAS, 2018). However, the PoD uses otherwise meaningless pseudo-random numbers to determine ledger recording privilege among participants, which is not consistent with the concept of utilizing resources only for meaningful and productive endeavors. Moreover, due to the limitation of design, the PoD cannot achieve the level of efficiency required by the AIBC.

### 5.2.5 PoA (Proof of Authority) Identity Proof Consensus

The PoA is similar to the PoS (VET, 2018). However, unlike the POS, the PoA nodes are not required to hold assets to compete for ledger recorder privilege, rather, they are required to be known and verified identities. This means that nodes are not motivated to act in their own interest. The PoA is cheaper, more secure and offers higher TPS than the PoS.

### 5.2.6 BFT (Byzantine Fault Tolerance) Distributed Consistency Consensus and DBFT

The BFT provides $F = \lfloor (N-1)/3 \rfloor$ fault tolerance. The possible solution to the Byzantine problem is that consistency can be achieved in the case of $N \geq 3F + 1$, where *N* is the total number of validators, and *F* is the number of faulty validators. After information is exchanged between the validators, each validator has a list of information obtained, and the information that exists in a 2/3 majority of validators prevails. The BFT advantage is that consensus can be reached efficiently with safety and stability (Lamport, Shostak and Pease, 1982; Driscoll et al., 2003).

A high-performance variant of the BFT, the PBFT (Practical BFT), can achieve a consensus delay of two to five seconds, which satisfies the real-time processing requirements of many commercial applications (Castro and Liskov, 2002). The PBFT's high consensus efficiency enables it to meet high-frequency trading needs.

The disadvantages of the BFT are that, when one third or more of the validators stop working, the system will not be able to provide services; and that when one third or more of the validators behave maliciously and all nodes are divided into two isolated islands by chance, the malicious validators can fork the system, though they will leave cryptographic evidence behind. The decentralization level of the BFT is not as high as the other consensuses, thus it is more suitable for multi-centered application scenarios.

The DBFT is to select the validators by their stake in the ecosystem, and the selected validators then reach consensus through the BFT algorithm (NEO, 2018). The relationship between DBFT and the BFT is similar to the relationship between DPoS and PoS. The DBFT has many improvements over the BFT. It improves the BFT's client/service architecture to a peer-node mode suitable for P2P networks. It evolves from static consensus to dynamic consensus that validators can dynamically enter and exit. It incorporates a voting mechanism based on the validators' stakes for ledger recording. It also introduces the usage of a digital certificate, which resolves the issue of validator identity authentication.

The DBFT has many desirable features, such as specialized bookkeepers, tolerance of any type of error, and no bifurcation. Just as with the BFT, when one third or more of the validators behave maliciously and all nodes are divided into two isolated islands by chance, the malicious validators can fork the system, though they will leave cryptographic evidence behind.

#### 5.2.7 DABFT – An Adaptive Approach

Thus, in view of the advantages and disadvantages of the existing consensus algorithms, we conclude that, although some of them offer useful features, none of them alone can fully meet the AIBC goals of efficiency, fairness, and legitimacy.

We thus propose the DABFT, which combines some of the best features of the existing consensus algorithms. Conceptually, the DABFT implements certain PoS features to strengthen the legitimacy of the PoI, and certain PoI features to improve the fairness of PoS. It also improves the PoD's election mechanism with the BFT algorithm.

In addition, the DABFT is further extended by a feature of adaptiveness. The DABFT is a delegated mechanism with a higher level of efficiency and is essentially a more flexible DBFT that is capable of selecting BFT flavors most suitable for particular (and parallel) tasks on the fly. The adaptiveness is achieved by deep learning techniques, that real-time choices of consensus algorithms for new tasks are inferred from trained models of previous tasks.

Therefore, the DABFT is the perfect tool to build the efficient, legit and fair AIBC ecosystem that conducts only meaningful and productive activities.

### 5.3 Design of DABFT Algorithm

#### 5.3.1 New Block Generation

Upon the release of a new task, a subset of super nodes that are most relevant to the task is selected as the representatives (task validators), who then elect among themselves a single task handler responsible for managing the task. The task handler then selects a number of resource nodes that are the most relevant to the task, and distribute the task to them. Upon successful release of the new task, the task handler proposes a new block that is then validated by the task validators. A new block is thus born.

Because of the "rule of relevancy," it is highly likely that each new task is assigned a completely different set of task validators and task handler. However, once the task handler and validators are selected, they manage the task from inception to completion (from the root block to the final block of the side chain). Therefore, there is no need for the periodical system-wide reelection of representatives. The key benefit of this arrangement is that no dynasty management is required, which reduces the system's complexity and improves its efficiency.

The real-time selection of task validators and handler for a new task based on the "rule of relevancy" means the DABFT has a built-in "dynamic sharding" feature, which will be explained in a later subsection.

### 5.3.2 Consensus Building Process

After a task handler proposes a new block, the task validators participate in a round of BFT voting to determine the legitimacy of the block.

At present, none of the mainstream BFT algorithms is optimal for all tasks. The DABFT utilizes a set of effectiveness evaluation algorithms through AI based deep learning to determine the optimal BFT mode for the task at hand. The flavors of BFT algorithms for the DABFT to choose from include, but not limited to, DBFT and PBFT (Practical BFT), as well as Q/U (Abd-El-Malek, 2005), HQ (Cowling, 2006), Zyzzyva (Kotla et al., 2009), Quorum (Guerraoui, 2009), Chain (Guerraoui, 2009), Ring (Guerraoui, 2011), and RBFT (Redundant BFT) (Aublin, Mokhtar and Quéma, 2013), etc. Figure 5.1 shows the consensus process for several mainstream BFT algorithms.

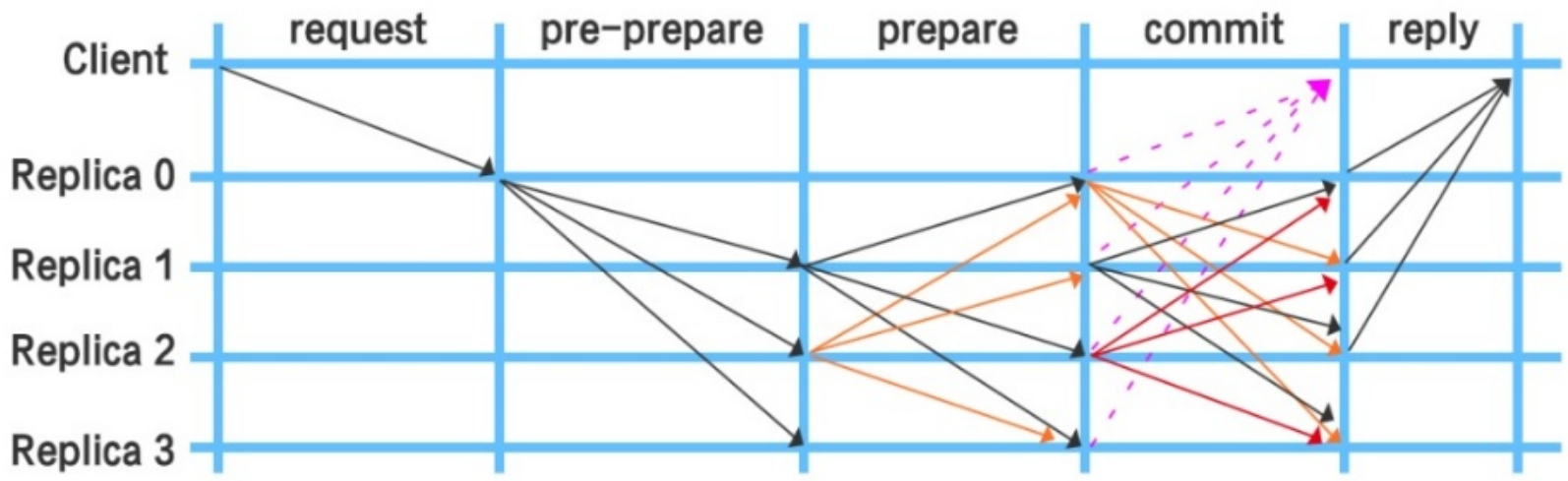

( a ) PBFT

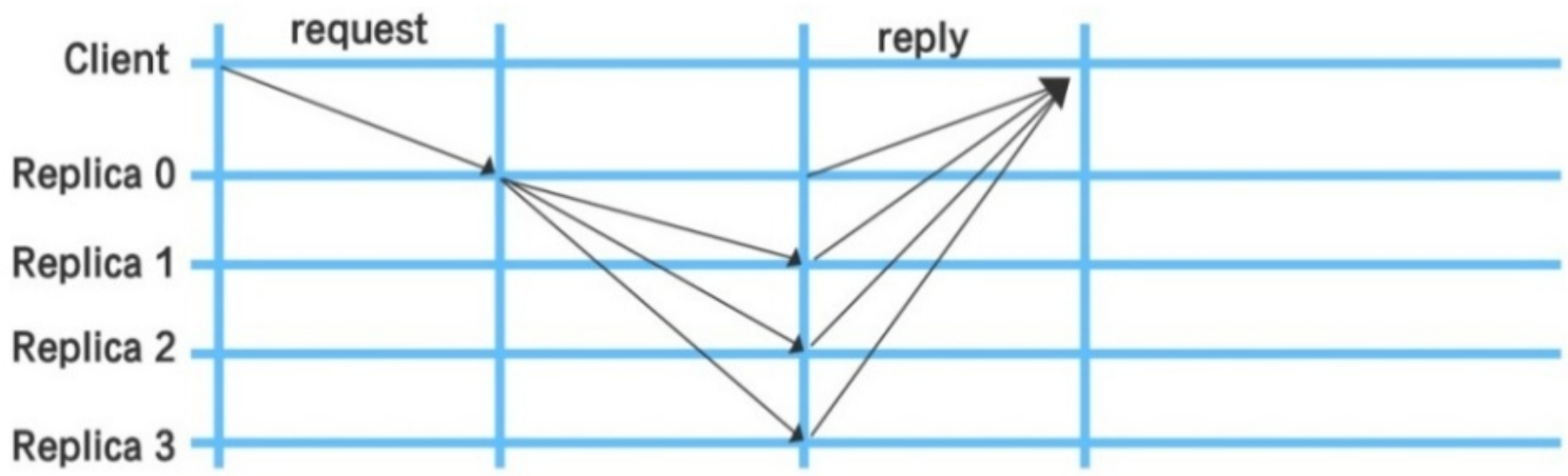

( b ) Zyzzyva

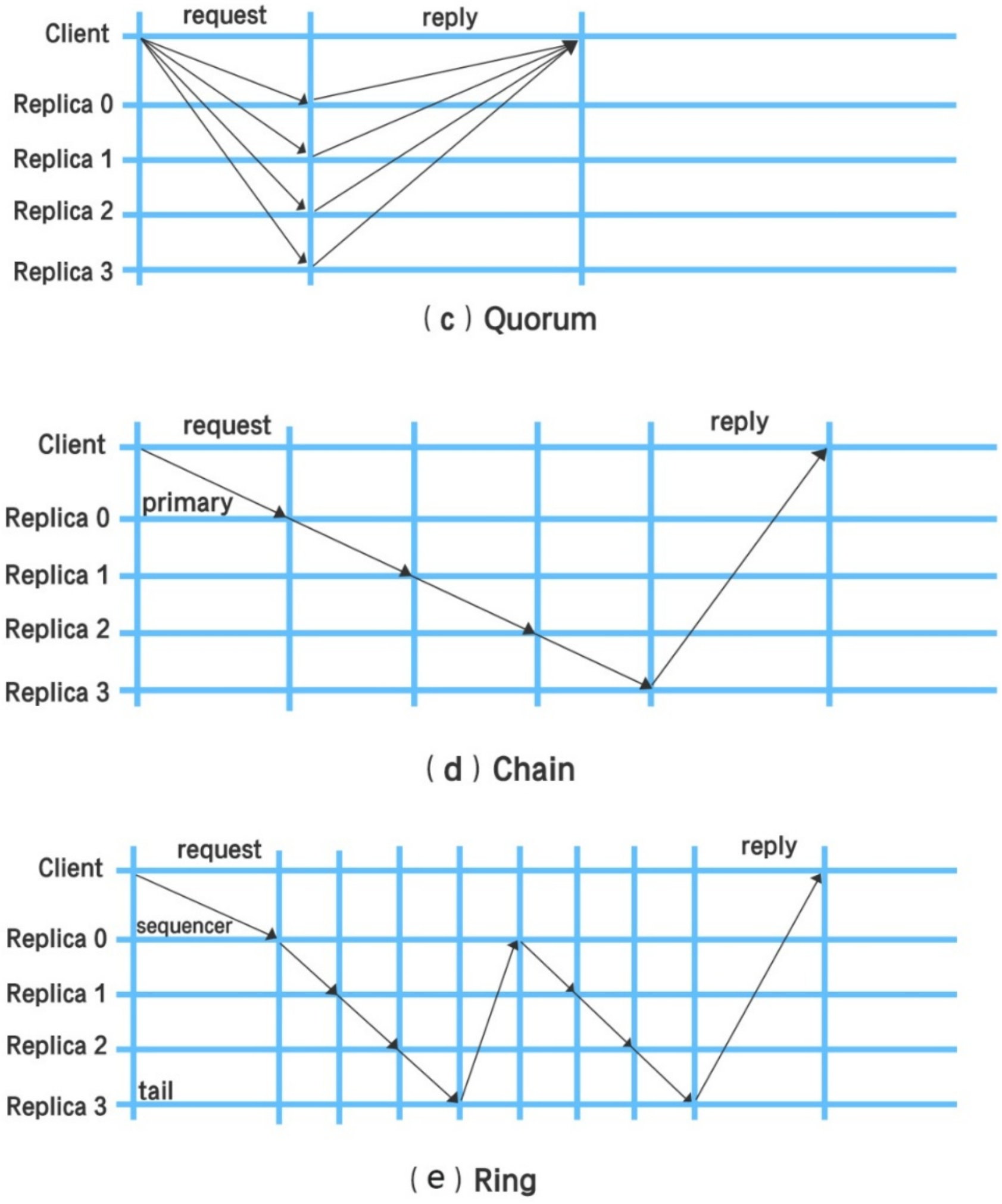

**Figure 5.1 – Mainstream BFT algorithm consensus process**

Through the machine learning prediction, DABFT dynamically switches the system to the optimal BFT consensus of the present task. The state of a modular monitoring system (number of clients, errors, message sizes, etc.) triggers this process, which is determined by the user preference of the Byzantine protocol and calculated score between the system protocol and its system performance matrix.

The DABFT improves upon the ADAPT (Bahsoun, Guerraoui, and Shoker, 2015) and is similar to it in several ways. Like the ADAPT, the DABFT is a modular design and consists of three important modules: BFT System (BFTS), Event System (ES), and Quality Control System (QCS). The BFTS is essentially an algorithms engine that modularizes the aforementioned BFT algorithms. The ES collects factors that have a significant impact on performance and security in the system, such as a number of terminals, requests, sizes, etc., and sends task information to the QCS. The QCS drives the system through either static (Shoker and Bahsoun, 2013), dynamic, or heuristics mode, and evaluates a set of Key Performance Indicators (KPI) and Key Characteristics Indicator (KCI) to select the optimal BFT flavor for the task at hand.

The QCS computes the evaluation scores of the competing BFT protocols for a particular task and then selects the protocol with the highest score. For a given task $t$ and protocol $p_i \in BFTS$ that has an evaluation score $E_{i,t}$ $(element\ of\ Matrix\ E)$, the best protocol $p_t$ is given as:

$$p_t = p_i, \ s.t.\ E_{i,t} = \max_{1 \le j \le n} E_{j,t} \tag{5.1a}$$

$$where \begin{cases} E = C \circ P \\ C = \left\lfloor \frac{1}{a} \cdot (A \dot{\vee} (e_n - U)) \right\rfloor \\ P = B^{\pm} \cdot (V \circ W) \end{cases} \tag{5.1b}$$

Where $C$ is the KCI matrix and $P$ the KPI matrix; matrix $A$ represents the profiles (i.e., the KCIs) of the protocols; Column matrix $U$ represents the KCI user preferences (i.e., the weights); column matrix $e_n$ is a unit matrix used to invert the values of the matrix $U$ to $-U$. The use of $1/a$ within the integer value operator $\lfloor\ \ \rfloor$ rules out protocols not matching all user preferences in matrix $U$. Matrix $B$ represents KPIs of the protocols, one protocol per row. Column matrix $V$ represents the KPI user-defined weights for evaluations. Column matrix $W$ is used in the heuristic mode only, with the same constraints as matrix $V$. The operator "∘" represents Hadamard multiplication, and the operator "$\dot{\vee}$" represents Boolean multiplication.

There is one major shortcoming in the ADAPT design. The ADAPT employs the Support Vector Regression (SVR) method (Smola and Schölkopf, 2004) with a five-fold cross-validation to predict the KPI parameters for elements in matrix $B$. There are six fields in the dataset: number of clients, request size, response size, throughput, latency, and capacity. While the methodology is useful in Byzantine fault tolerance settings in and of itself, it is not designed for the highly complex blockchain application scenarios in which there are many interactions between participants, thus it is not particularly effective for them. For example, in the AIBC context, at any given time there are multiple tasks (handled by different handlers) that compete for resources. As such, the ever-increasing number of tasks and interactions between them affect the key KPI parameters (throughput, latency, and capacity) for individual tasks continuously (time series), for the purpose of achieving the best performance on the system-level. As such, it is necessary to introduce a mechanism that incorporates time-varying conditional correlations across tasks in order to adjust the KPI parameters on the fly. What sets the DABFT apart from the ADAPT is that the DABFT has such a function built in.

The DABFT implements the time-varying conditional correlation mechanism in the QCS. First of all, for task $t$, the QCS trains on the existing data to come up with the initial matrix $\hat{B}_t$ (basically matrix $B$ in the ADAPT, but specifically for task $t$). It then calculates a residual matrix $E_t$ as follows[1]:

$$E_t = \hat{B}_t - B_t \tag{5.2}$$

Where $B_t$ is the "real" KPI parameter matrix derived from empirical tests.

The specification with time-varying multi-dimensional correlation matrix for task $t$ is thus given as[2]:

$$\begin{aligned}
&\mathrm{E}_t \mid \Psi_{t-1} \sim N(0, \Omega_t = \mathrm{H}_t \mathrm{P}_t \mathrm{H}_t) \\
&\mathrm{H}_t^2 = \mathrm{H}_0^2 + \mathrm{K}\mathrm{E}_{t-1}\mathrm{E}_{t-1}^T + \Lambda \mathrm{H}_{t-1}^2 \\
&\mathrm{P}_t = O_t^* O_t O_t^* \\
&\Xi_t = \mathrm{H}_t^{-1}\mathrm{E}_t \\
&O_t = (1-a-b)\overline{O} + a\Xi_{t-1}\Xi_{t-1}^T + bO_{t-1} \\
&a + b < 1
\end{aligned} \tag{5.3}$$

Where:

1. $\mathrm{E}_t$ is the conditional residual vector at time $t$ given the previous state $\Psi_{t-1}$.
2. $\Omega_t$ is the conditional covariance matrix of $\mathrm{E}_t$.

---

[1] The $\hat{B}_t$ and $B_t$ are full matrices made of row vectors for individual BFT flavors, while $E_t$ is actually a column matrix. The mathematical representation in this subsection is simplified just to illustrate the analysis process without losing a "high-level" accuracy.

[2] Essentially, this is a Dynamic Conditional Correlation (DCC) for multivariate time-series analysis with a DCC(1,1) specification (Engle and Sheppard, 2001; Engle, 2002).

3. $\mathrm{P}_t$ is the conditional correlation matrix of $\mathrm{E}_t$.
4. $\mathrm{H}_t$ is the normalization matrix for $\mathrm{P}_t$.
5. $\mathrm{K}$ and $\Lambda$ are diagonal coefficient matrices for $\mathrm{H}_t$.
6. $\Xi_t$ is the standardized residue vector of $\mathrm{E}_t$.
7. $O_t$ and $O_t^*$ are estimator matrices for $\mathrm{P}_t$.
8. $\overline{O}$ is the unconditional correlation matrix of $\mathrm{E}_t$.

It's worth mentioning that Equations (5.2) and (5.3) only propagate from task $t$ back to task $t-1$ for the purpose of reducing computation complexity.

Finally, the predicted KPI matrix for task $t$, $\bar{B}_t$, is given as:

$$\bar{B}_t = \hat{B}_t + \Omega_t \tag{5.4}$$

From this point and onward, DABFT is similar to the ADAPT, and proceeds to select the BFT protocol with the evaluation highest score based on Equations (5.1a) and (5.1b). For any BFT choice, the DABFT provides fault tolerance for $F = \lfloor (N-1)/3 \rfloor$ for a consensus set consisting of *N* task validators. This fault tolerance includes security and availability and is resistant to general and Byzantine faults in any network environment. The DABFT offers deterministic finality, thus a confirmation is a final confirmation, the chain cannot be forked, and the transactions cannot be revoked or rolled back.

Under the DABFT consensus mechanism, it is estimated that a block is generated every 0.1 to 0.5 seconds. The system has a sustainable transaction throughput of 30,000 TPS, and with proper optimization, has a potential to achieve 1,000,000 TPS, making the AIBC ecosystem capable of supporting high-frequency large-scale commercial applications.

The DABFT has an option to incorporate digital identification technology for the AIBC to be real name based, making it possible to freeze, revoke, inherit, retrieve, and transfer assets under judicial decisions. This feature makes the issuance of financial products with compliance requirement possible.

### 5.3.3 Fork Selection

The DABFT selects the authority chain for each task with a block score at each block height. Under the principle of fairness and legitimacy, the forked chain of blocks with the highest economic value is selected to join the authority chain. The economic value of a forked chain is the sum of the economic value of the forked block and the descendants of that block. This is achievable because all tasks are tracked by their corresponding side chain blocks that will eventually reach finality.

### 5.3.4 Voting Rules

In order to defend against malicious attacks to the consensus process, the DABFT borrows Casper's concept of minimum penalty mechanism to constrain task validators' behavior. The voting process follows the following basic rules:

1. The consensus process of a single block has a strict sequence. Only after the total number of votes in the first stage reaches 2/3 majority, can the next stage of consensus start.
2. The consensus of a subsequent block does not need to wait until the consensus of the current block is concluded. The consensuses of multiple blocks can be concurrent, however not completely out of order. Generally, after the consensus of the current block is 2/3 completed, the consensus of a subsequent block can start.

### 5.3.5 Incentive Analysis

The task validators (including the task handler) participating in the DABFT of a task receive rewards in the form of CFTX tokens according to the DPoEV incentive consensus. The total number of tokens awarded to the task validators is a percentage of the overall number of tokens allocated to the task and is shared by all participating task validators and handler. The number of tokens awarded to the task handler and each task validator is determined by its contribution to the completion of the task. These numbers are dynamically determined by the DPoEV, particularly its EVG engine.

### 5.3.6 Cheating Analysis

There are several attacks of particular interest in distributed consensus, and three of the most analyzed ones are double spending attack, short-range attack and 51% attack. In the DPoEV-DABFT two consensus AIBC ecosystem, by design, none of the attacks have a chance to succeed.

A double spending attack happens when a malicious node tries to initiate the same tokens through two transactions to two distinguished destinations. In a delegated validation regime (e.g., DPoS or DBFT), for such an attack to succeed, the malicious node must first become a validator through the election (with deposit paid) and then bribe at least one-third of other validators in order for both transactions to reach finality. It is impossible to succeed in double spending in the DPoEV-DABFT two consensus AIBC ecosystem. The reasons are that the validators (super nodes) are chosen by their relevancy to tasks but not their deposits; that the validators are not allowed to initiate tasks; and that the validators are rewarded based on their levels of contribution, not by other validators. Essentially, conditions for the double spending attack to occur do not exist.

A short-range attack is initiated by a malicious node that fakes a chain (A-chain) to replace the legitimate chain (B-chain) when the H+1 block has not expired. In a delegated regime for this attack to be successful, the attacker needs to bribe the validators in order to make the block A1 score higher than B1. Thus, essentially, the short-range attack is very much like a double-spending attack at the A1/B1 block level, which has no chance to succeed for the same reason that makes the double-spending attack futile.

In the PoW, a 51% attack requires a malicious node to own 51% of the total computing power in the system, in the PoS 51% of the deposit, and in the PoD 51% of the certified accounts. In the DPoEV-DABFT two consensus AIBC ecosystem, restrained by the economic model, there is no possibility for any node to own more than 51% of the economic value. More importantly, since the validators are not allowed to initiate tasks (thus transactions), a validator with bad intention must bribe its compatriots to even launch such an attack. However, the validators are rewarded based on their levels of contribution, not by other validators. Essentially, conditions for the 51% attack to occur do not exist either.

### 5.3.7 Dynamic Sharding

One of the challenges the mainstream blockchains face is scalability, which is key to performance improvement. Ethereum seeks to resolve the scalability issue with the so-called sharding approach, in which a shard is essentially "an isolated island" (Blockgeeks, 2018; Sharding 2018). The DABFT, by design, has a built-in dynamic sharding feature.

First of all, the AIBC ecosystem is a 2D BlockCloud with super nodes that track the status of tasks through side chains. Once a task is initiated, a set of task validators are then selected according to the "rule of relevancy." A task handler is then chosen among the task validators to handle the task. The task handler and validators manage the task from the beginning to the end with no dynasty change. Thus, effectively, from the task's perspective, the task validators form a shard that is responsible for managing it, with the task handler being its leader.

In addition, due to the "rule of relevancy," it is highly likely that each new task is assigned a different set of task validators from the previous task, although overlapping is possible, especially when the number of super nodes is small. Therefore, once a task is completed and its associated task reaches finality, its shard dissolves automatically. Therefore, in the AIBC, no periodic "re-sharding" is necessary. Such fluidity affords the AIBC a "dynamic" sharding feature.

The dynamic sharding feature makes the so-called single-shard takeover attack against the AIBC impossible to succeed. First off, shards are directly formed by tasks in a highly random fashion due to the unpredictable nature of the "rule of relevancy." Second, shards have very short lifespans as they only last till tasks are completed. Practically, malicious nodes never have a chance to launch attacks.

The AIBC also maintains a 1D "main chain" at each super node, with the blocks of side chains of shards intertwined. A Merkle tree structure of the 1D blockchain makes it topologically identical to the 2D BlockCloud.

# 6 AIBC Operations

## 6.1 Task driven

The AIBC ecosystem is task driven. Only tasking nodes can generate tasks. A task is a collection of smart contracts. Each smart contract, called assignment, is an independent subtask and is the smallest (indivisible) executable unit that, if executed successfully by a resource node, generates an acknowledgment from that node, which is then propagated back to a super node (the task handler). A tasking node needs to have sufficient amount of CFTX tokens to purchase services rendered by the service nodes. A P2P (Peer-to-Peer) token only transaction is the smallest possible task with only one assignment (smart contract) embedded.

## 6.2 Task structure

The AIBC adopts the concept of "blocks track task" and "transactions track assignment." That, when a tasking node initiates a task sends it out, an elected super node (the task handler) proposes a block based on the task, which is then validated by a group of super nodes (the task validators). After the validation is complete, the task becomes the root block for itself. The side chain attached to the task handler then grows with additional blocks, with each block tracking the progress of the task ("blocks track task"). Within a block, a transaction reflects the state of an assignment at the moment when the block is time-stamped, that is, a series of transactions tracks the progress of an assignment from initiation to completion ("transactions track assignment"). When the task is completed or abandoned, its "final" block is then added to the chain, at which point the task is closed permanently, without any probability to be reopened.

While the number of blocks per task can vary because of assignment multiplicity, the assignment length (number of transactions per assignment), or state of assignment, is always limited to four: assignment initiation (tasking node and task handler), assignment acceptance and acknowledgment (task handler and resource nodes), assignment completion and acknowledgement (resource nodes and task handler), and assignment close (task handler and tasking node).

A block is thus a data structure with a block/transaction format that contains, but is not limited to, the following fields:

- Header: block header
  - Height: block height of the task
  - ParentHash: parent block hash
  - Ts: timestamp
  - Tnad: tasking node address
  - Thad: task handler (elected super node) address
  - Epoch: the consensus age of the block
  - BTimer: timer for block (task) expiration
  - AssignNum: number of assignments in the task
  - StateRoot: state root hash
  - TxsRoot: transaction root hash
  - ReceiptsRoot: transaction receipt hash
- CFTX index: amount of wealth (number of CFTX tokens) for this task, determined by the DPoEV dynamically
  - GlobalWealth: total wealth in the ecosystem
  - TaskWealth: wealth entitled to this task
  - THWealth: task handler wealth
  - TRelevancy: task relevancy index
- Transactions: transaction data tracking assignments (including multiple transactions)
  - AssignType: whether the assignment is a request for computing power, stored data or NULL (for token only transactions)
  - AssignID: assignment ID to be tracked
  - AssignWealth: a wealth of the assignment this transaction tracks
  - TranState: Transaction state
  - From: transaction sender address
  - To: transaction receiver address

- Value: transfer amount, dynamically determined
  - Data: transaction payload, smart contracts for the resource nodes to execute
  - Signature: transaction signature
  - Gas: gas limit, dynamically determined
  - GasPrice: gas unit price, dynamically determined
  - Nonce: the uniqueness of transactions
  - TTimer: timer for the transaction (assignment state) expiration
- Votes: Prepare and Commit Votes (including multiple), used in DABFT consensus algorithm
  - TypeBFT: type of Byzantine Fault Tolerance algorithm for this block, determined dynamically by DABFT
  - From: voter address
  - VoteHash: hash of the block voted for
  - Hv: the height of the block voted for
  - Hvs: the height of an ancestral block of the block voted for
  - VoteType: voting type, Prepare or Commit
  - Signature: vote signature
- Version Code: The version code for protocol update
  - Hash: hash of the version code
  - Code: the bytecode of the version code
  - IniBlock: the initial block number of the current version
  - Signature: signature (sign by the developer community)
  - Version: the version code number, upgraded incrementally
  - Nonce: the uniqueness of protocol code

## 6.3 AIBC Nodes

The AIBC is a hierarchical virtual cloud with three types of nodes: super nodes, tasking nodes, and resource nodes. The resource nodes are further classified as computing nodes and storage nodes. Thus, effectively and from a task perspective, there are four types of nodes: super nodes, tasking nodes, computing nodes and storage nodes. Other than the super nodes, the tasking nodes and resource nodes can physically or virtually be collocated or coincide. Their roles and responsibilities are specified in this section.

### 6.3.1 Super Nodes

The super nodes reside in the fundamental layer of the AIBC, and are the "real" blockchain nodes. Only super nodes are allowed to approve and broadcast tasks to resource nodes. The super nodes follow the DABFT distributed consensus strictly when they carry out the following actions:

1. Task validators and handler selection: All tasking nodes are entitled to submitting tasks. Once a tasking node generates and broadcasts a task request to the AIBC, the next step is for the super nodes to elect a task handler among themselves to manage the task. First, a subset of super nodes, called a task validator set, is chosen. The key selection criteria for the set are relevancy (that the super nodes are most relevant to the task, or the "rule of relevancy" based on the ERR rankings) and then wealth level (that only super nodes with the lowest levels of wealth are chosen if there is a higher number of relevant nodes than what is required for the task, or "rule of wealth"). Then, a task validator with the lowest level of wealth is selected as the task handler or the "block miner" that serves as the block proposer and ledger recorder. Once a task handler is selected, it retains the right to proposing all subsequent blocks pertaining to the task until the task is closed permanently.
2. Block proposition and validation: The task handler reviews the task and checks its intention. If the structure of the task is in compliance, the task handler registers the task, approves it and then proposes an initial task block, the root block. The AIBC adopts a blocks-track-task concept, that, a task submitted by a tasking node is essentially the root block that the task handler will propose, providing that everything about the task is in compliance with the block data structure. The task root block is then validated by the task validators.

3. Task dissemination and assignment: After a task (root block) is validated, the task handler automatically disseminates it into a series of executable smart contracts, based on the information provided in the header of the task data structure. Each smart contract, called assignment, is an independent subtask and is the smallest (indivisible) executable unit. Each smart contract is then distributed (assigned) to a set of resource nodes that are most suitable to execute it ("rule of relevancy"). The task handler determines the number of resource nodes required to complete the assignment. If there are more resource nodes than what is required of the assignment, the task handler selects a subset of resource nodes with the lowest levels of wealth ("rule of wealth"). The choice of this set of resource nodes is then validated by the task validators. The task handler then records the task dissemination and assignment state (with states of other assignments) in a subsequent block, which is again validated by the task validators.
4. Assignment acknowledgment and reassignment: The resource nodes that have received and accepted the assignment send an acknowledgment back to the task handler. If within a given time period, the task handler has not received a predetermined number of acceptance acknowledgments, it distributes the assignment to an alternative group of resource nodes (validated by task validators), without releasing the resource nodes that have already accepted the assignment in the previous round(s). The process repeats until the task handler eventually receives a predetermined number of acceptance acknowledgments, or an assignment timer[3] expires. Upon expiration of the timer, the task handler makes an inquiry to the tasking node that initiated the task, which then determines if that assignment will be distributed again (to yet a different set of resource nodes) or abandoned. The task handler then updates the state of that assignment and records the assignment acknowledgment and/or reassignment in a subsequent block, which is then validated by the task validators.

[3] The assignment timer can be set differently for assignments with different time criticalness.

5. Assignment output (result) collection and submission: If sufficient resource nodes have accepted an assignment, the task handler instructs them to execute the smart contract. For a resource node, either it successfully executes the contract and produces a legitimate result, or it fails and produces a FALSE output. In case only FALSE outputs are produced for an assignment, the task handler makes an inquiry to the tasking node, which then determines if the assignment will be redistributed or abandoned. If at least one legitimate result is produced, the task handler packs the result with all other relevant information and submits the package back to the tasking node. The task handler then updates the state of that assignment (with states of other assignments) and records the assignment acknowledgment and/or reassignment in a subsequent block, which is then validated by the task validators.
6. Task wrap-up and bookkeeping: The task handler keeps collecting assignment results until it receives responses for all assignments of the original task, be these responses legitimate, FALSE, or timeout. It then aggregates the assignment responses into a task completion package and sends the package back to the tasking node. The task handler then records the task wrap-up information and broadcasts the task completion record to other task validators. The task handler then updates the task completion status in a subsequent block, which is then validated by the task validators. Once the task handler receives an acknowledgment from the tasking node that it has received the task completion package, the task is closed and no longer accepts any update, and this subsequent block becomes the final block of the task. The task handler then relinquishes its position.
7. All participating super nodes are awarded CFTX tokens. The total number of tokens for each task, to be shared by all participating nodes (both super nodes and resource nodes), is determined by the DPoEV incentive consensus. The number of tokens awarded to all participating super nodes (task handler and validators) is a percentage dynamically determined by the DPoEV. The number of tokens awarded to each participating super node depends on its contribution to task completion.

### 6.3.2 Tasking Nodes

The tasking nodes are critical in the AIBC ecosystem. They are the nodes with needs for resources such as computing power and storage space, thus it is their responsibility to initiate tasks and drive the wealth generation process. A tasking node can also be a resource node and vice versa.

The main responsibilities of a tasking node are as follow:

1. Tasks generation: The very reason for the AIBC to exist in the first place is that there are needs for computing power and storage space, and the needs reside in the tasking nodes. All tasking nodes are entitled to generating tasks for its computing and storage needs. A task, from the tasking node's perspective, is a request for information that can be further utilized by its local business logic. For example, a tasking node that runs quantitative trading algorithms can request computing nodes to predict the next day's return of a stock; once the computing nodes complete stock return predictions and submit the predictions back to the tasking node, it then feeds the predictions to a variety of quantitative models to produce investment portfolios. Furthermore, it is the tasking node's responsibility to assemble the task (block) data structure with a collection of assignments (smart contracts), with different methodology and dataset for each assignment. Using the same example, when a tasking node initiates a task, which is a request to predict the next day's return of a stock, it provides different algorithms and datasets to different computing nodes (though the task handler), expecting them to return different predictions in order to construct distinguished investment portfolios. Each combination of algorithm and dataset constitutes an assignment within a task.

2. Assignment acknowledgment reception: If a tasking node gets a confirmation from the task handler that an assignment within a task has been accepted by a predetermined number of resource nodes, it starts anticipating the eventual arrival of the assignment result. After a sufficient number of assignments (of a task) has been accepted, it instructs its local computing unit to schedule the execution of the local business logic. If, however, a sufficient number of assignments has not been successfully acknowledged before a task acknowledgment timer expires, the tasking node must decide whether it needs to resend the task, or abandon it.
3. Assignment completion reception: A tasking node then keeps receiving assignment results from the resource nodes (through the task handler), and making a determination whether it has received a sufficient number (above a predetermined threshold) of assignment results to execute its local business logic before a task execution timer expires. If it does, it starts executing the logic. If it does not, it must decide whether it needs to resend the task or abandon it. It then sends a status update to the task handler.
4. Task wrap-up: Once the tasking node receives the task completion package from the task handler, it acknowledges the task handler, which in turn closes the task permanently and proposes the final block of the task. If, after a task timer expires and the tasking node still has not received the task completion package, it must decide whether it needs to resend the task, or abandon it.
5. If a tasking node wishes to generate a task, it must have sufficient CFTX tokens to "purchase" the computing power and/or storage space from the resource nodes, as well as block proposing and ledger recording services provided by the super nodes. The required number of tokens for a particular task is determined by the DPoEV consensus according to the economic value added to the ecosystem.

### 6.3.3 Computing Nodes

When a computing node takes on an assignment, essentially it becomes a task miner, of which the main responsibilities are as follow:

1. Assignment acceptance: The main service that a computing node performs is accepting and completing assignments. When a computing node gets an assignment from the task handler, it makes a determination whether it has all the needed information (model, data, etc.) and sufficient computing power to complete the assignment. If it does, it sends an acknowledgment to the task handler that it has accepted the assignment. If it does not, it sends an acknowledgment to the super nodes that it has rejected the assignment, or it needs more information (e.g., model, data, etc.).
2. Assignment completion. After a computing node accepts an assignment, it executes the associated smart contract. If it cannot successfully execute the smart contract, or an assignment timer has expired, it sends a FALSE message to the super nodes. If it successfully executes the smart contract, it sends the assignment output (results) back to the task handler.
3. All participating computing nodes are awarded CFTX tokens. The total number of tokens for each completed assignment, determined by the DPoEV consensus, directly reflects its contribution to the ecosystem and is shared by all participating computing nodes that have accepted that assignment. The number of tokens awarded to each participating computing node depends on the assignment level of difficulty, model(s) the node evokes, data it consumes, and quality of the results.

### 6.3.4 Storage Nodes

When a tasking node initiates a task, it is quite likely that a majority of the computing nodes do not have the needed data to complete their perspective assignments. Therefore, some of the assignments in the task are requests for data, which reside on certain storage nodes. It is the task handler's responsibility to forward a data assignment (as opposed to computing assignment) to a number of storage nodes that may have what is required. When a storage node receives a request (assignment) for data, it essentially becomes a data miner, and it performs the following:

1. Assignment acceptance: The main service that a storage node performs is serving data needs for computing nodes. When a storage node gets an assignment from the task handler, it makes a determination whether it has the needed information data (or a subset of it). If it does, it sends an acknowledgment to the task handler that it has accepted the assignment. If it does not, it sends an acknowledgment to the super nodes that it has rejected the assignment.
2. Assignment completion: After a storage node accepts an assignment, it executes the associated smart contract. If it cannot successfully execute the smart contract, or an assignment timer has expired, it sends a FALSE message to the super nodes. If it successfully executes the smart contract, it sends the assignment output (data) directly to the computing node that needs the data (not back to the task handler). It also sends an assignment complete acknowledgment to the task handler.
3. All participating storage nodes are awarded CFTX tokens. The total number of tokens for each completed assignment directly reflects its contribution to the ecosystem and is shared by all participating storage nodes that have accepted that assignment. The number of tokens awarded to each participating storage node depends on the assignment time criticalness, data amount, and data quality.

## 6.4 AIBC Working Principles

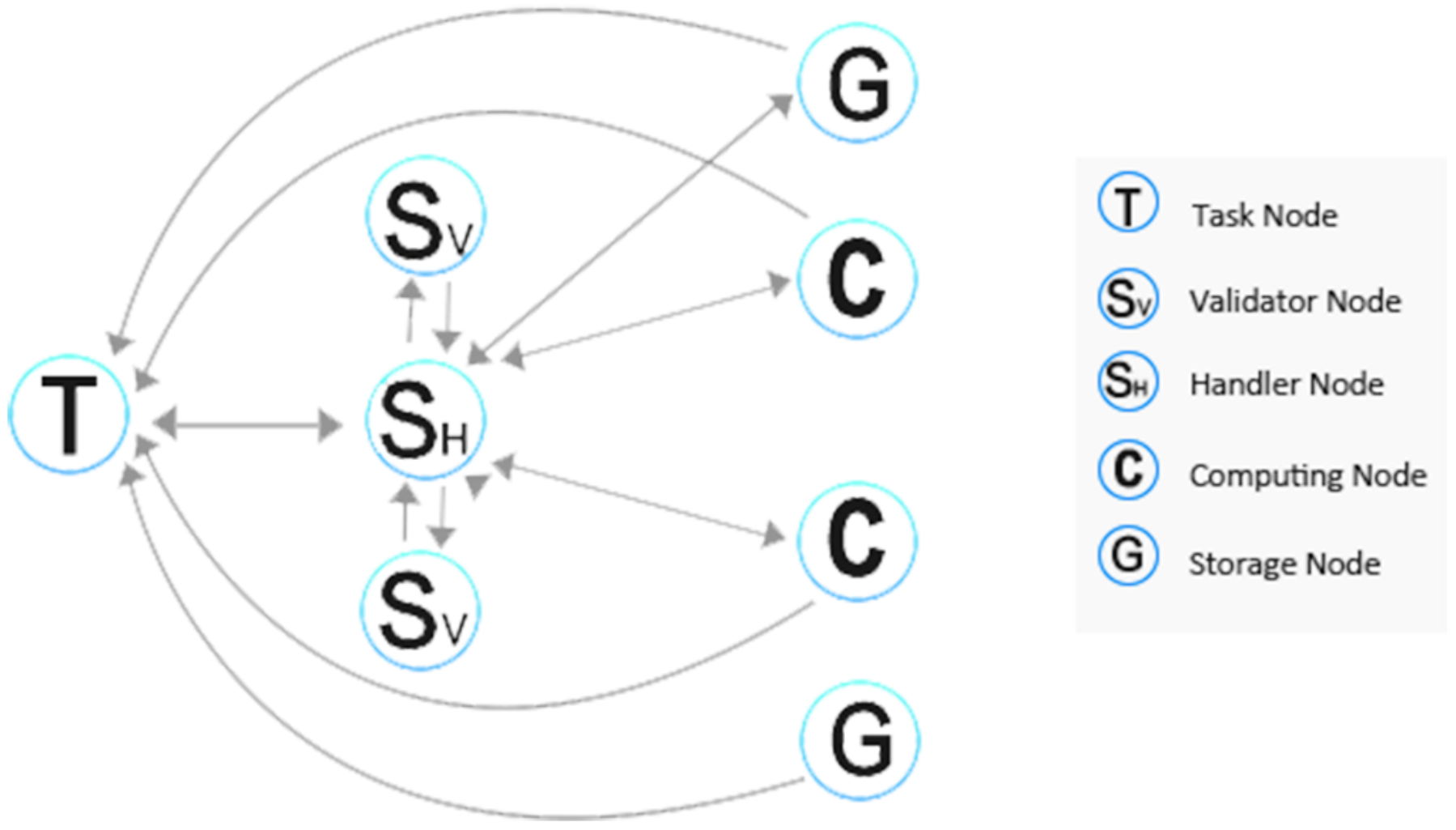

Figure 6.2 – AIBC Working Principles

## 6.5 AIBC Tasks Flow chart

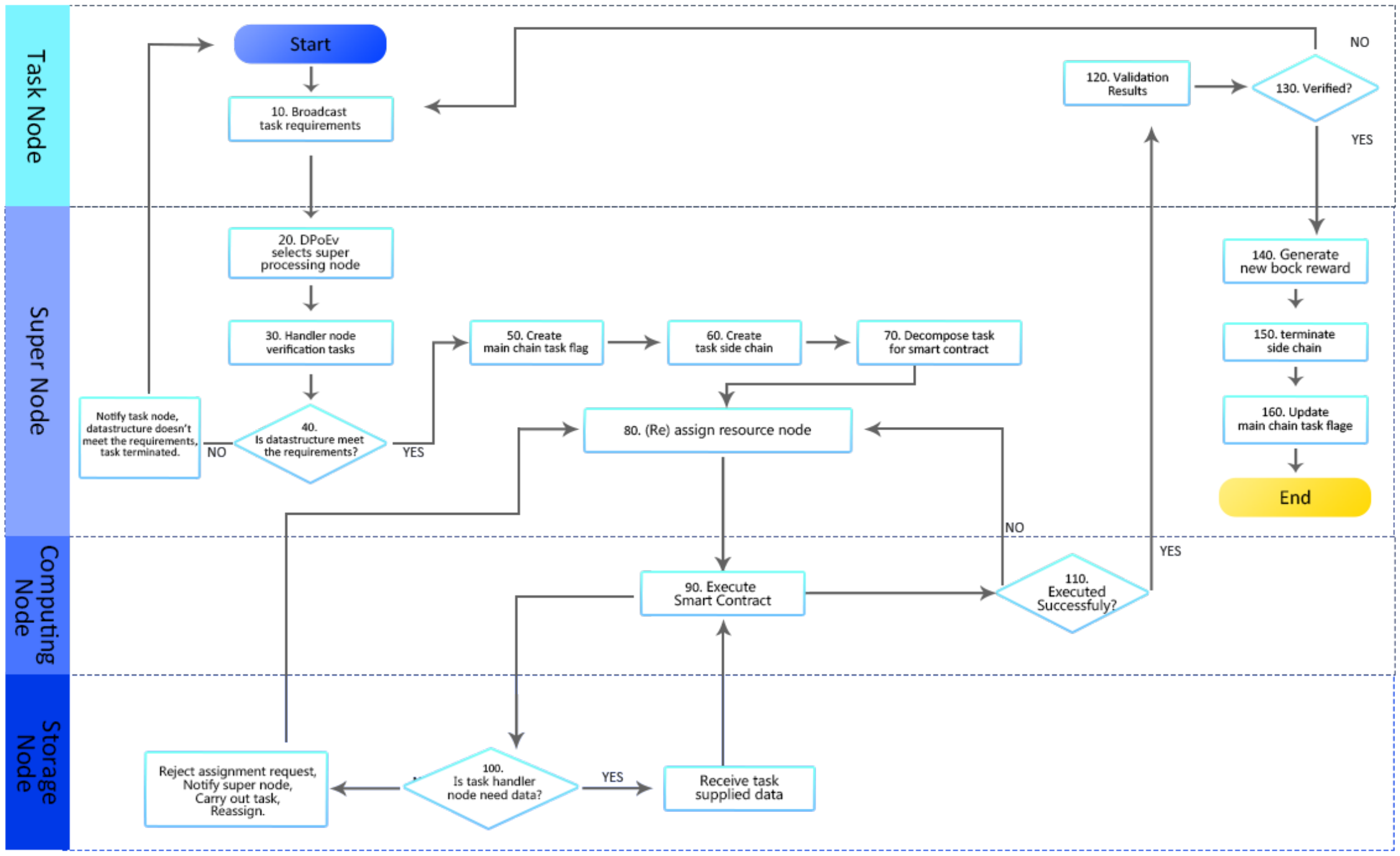

Figure 6.3 – AIBC Tasks Flowchart

# 7 AIBC Ecosystem

## 7.1 Dual-Token DSOL Platform

On the ecosystem layer, the AIBC is a “dual-token” platform that marks each decentralized application as a unique entity, yet provides a unified cross-platform value measure. AIBC offers a complete end-to-end distributed industry solution, Distributed Solution (DSOL)

In the AIBC ecosystem, each Distributed Solution (DSOL) is issued a number of its own distinguishable tokens, the DSOLxxxx (xxxx represents a number, not to be confused with the X in CFTX). The DSOLxxxx is a certifiable token, in other worlds each and every one of them is unique and individually trackable. The name “DSOLxxxx” is just a symbol that can have its own selected name (such as, say, AAAA).

The above concept can be illustrated by an example: Suppose that there are two DSOLs, DSOL1 and DSOL2, both of them are automated smart investments. DSOL1 issued 1,000 of its own tokens from DSOL1000 to DSOL1999 and DSOL2 issued DSOL2000 to DSOL2999. At the beginning, developers for DSOL1 and DSOL2 acquire separate and identifiable tokens DSOL1xxx and DSOL2xxx respectively for each of the two DSOLs to determine that DSOL has landed AIBC ecosystem with the assets of same initial value (for example DSOL1xxx and DSOL2xxx can be preset to the same initial value, say 100 CFTX, so DSOL1xxx=DSOL2xxx=100 CFTX, therefor the total value of DSOL1 and DSOL2 is 100,000 CFTX). Each DSOL, operates much like an independent micro-ecosystem within the AIBC ecosystem, accumulates wealth through value creation activities. For each DSOL founder or participant, incremental wealth can be created based on the DPoEV incentive consensus principle.

In our example, DSOL1 is an automated investment advisor that creates investment portfolio strategies. Each strategy created by DSOL1 yields an investment return forecast, which is then delivered to its users (investors) for them to build investment portfolios. These activities (strategy production, forecast deliver, investment results, etc.) collectively produce incremental knowledge in the DSOL1 micro-ecosystem, and mine additional CFTX tokens according to the DPoEV incentive consensus. Assuming that, at the beginning, DSOL1 is assessed an initial value of 100,000 CFTX (thus, DSOL1 = 100,000 CFTX), and after its first task, it creates an incremental knowledge valued at an additional 100 CFTX. Thus, after the completion of the first task, DSOL1 has a valuation of 100,100 CFTX (now DSOL1 = 100,100 CFTX).

At the meantime, another investment advisor DSOL2 is given the same task as DSOL1. However, the quality of the forecast produced by DSOL2 is inferior to the one produced by DSOL1, therefore, the return of the investment is lower. As a result, after the task is completed, it creates an incremental knowledge that is worth only 10 CFTX, thus, DSOL2 has a valuation of 100,010 CFTX (now DSOL2 = 100,010 CFTX).

As such, after completion of the first task, the values of DSOL1 and DSOL2 start to diverge, as now DSOL1 (= 100,100 CFTX) is worth more than DSOL2 (= 100,010 CFTX). The value of a DSOL reflects the real time state of wealth of its micro-ecosystem. As time progresses, the values of DSOL1 and DSOL2 may diverge even further. Essentially, the DPoEV incentive consensus guarantees that the DSOL produces higher amount and quality of knowledge will have higher valuation.

Since each DSOL contains 1,000 DSOLxxxx, the value of a DSOLxxxx can be one thousandth of the DSOL's value. For DSOL1, aft the completion of the first task, the 1,000 DSOL1xxx's may each be worth 100.1 CFTX (= 100,100/1,000). However, since each DSOL1xxx is unique and distinguishable, they can have different values. Indeed, it can occur, that DSOL1000 worth 1000 CFTX, and DSOL1999 10 CFTX, as long as the aggregated value of all DSOL1xxx's is equal to 100,100 CFTX.

The double-token DSOL ecosystem has two characteristics. First of all, on the AIBC ecosystem level, there are many decentralized applications, which are standalone yet interconnected micro-ecosystems that, essentially, can be regarded as companies in the AIBC "economy." Each DSOL is allowed only a finite number of tokens (DSOLxxxx), of which the aggregated value represents the total value of that DSOL. Each DSOLxxxx is unique, represents a fractional value of that DSOL and is not divisible. As such, a fractional ownership of DSOL can be transferred with the change of ownership of one or more DSOLxxxx token(s) but its functionality is still achievable. This design guarantees that a DSOL is always unique and cannot split into multiple DSOLs, that it represents only one application scenario in the AIBC ecosystem.

On the DSOL micro-ecosystem level, the value of a DSOL is measured by the amount of CFTX it contains. In our example, after the first task is completed, DSOL1 is worth 100,100 CFTX, thus, if DSOL1 was a company, its “book value” would be 100,100 CFTX. Therefore, the dual-token approach provides a unified measure of value (the CFTX token) across DSOLs, so that there can be fair comparisons of values among DSOLs.

An additional attribute is that, while each DSOL has a book value, its “market value,” which depends upon the quality and outlook of the DSOL itself, can be different. Again in our example, after the first task is completed, DSOL1 has a “book value” of 100,100 CFTX, and DSOL2 100,010 CFTX. From an investor’s perspective, they may regard that DSOL1’s superior performance as an indication that it will produce further superior performance in the future, thus they may price it with a market value that is higher than its book value, say, a 10% premium, or 110,110 CFTX (= 100,100 x 1.10 CFTX = 110,110 CFTX). On the other hand, they may assess DSOL2 with a 10% discount, or 90,009 CFTX (= 100,010 x 0.90 CFTX = 90,009 CFTX). Thus, DSOLs are incentivized to engage in productive activities.

## 7.2 Advantages of Dual-Token DSOL Platform

The dual-token approach allows the CFTX be used as the unified measure of value and transaction medium for the entire AIBC ecosystem, while at the same time enables the transfer of DSOL ownership on a whole sale level through auctions of DSOLxxxx tokens.

In the AIBC ecosystem, flexible design of DSOL-level securities is possible. In our example, if we were to make an analogy between the AIBC ecosystem and the financial world, token DSOL1000 can be a closed-end fund, of which the performance contains two parts: the performance of the underlying asset (CFTX), and the supply & demand of the fund (DSOLxxxx) itself. Token DSOL1001 can be an Asset Backed Security (ABS), of which the value depends upon the cash flows generated by DSOL1 assets, while token DSOL1002 can be a Credit Default Swap (CDS), of which the premium is derived from ratings of DSOL1 debts. The values of these DSOL1xxx tokens are measure by the amount of CFTX tokens they contain (book value), as well as their perceived quality and outlooks (market value).

The exchange of the DSOLxxxx tokens is done through digital asset auction platforms, whereas the exchange of the AIBC system token CFTX is done through digital asset exchanges.

## 7.3 System Implementation based on Dual-Tokens

For the first release of AIBC ecosystem, the dual-token DSOL mechanism will be built upon the Ethereum platform. The Ethereum platform offers the ERC721 token, each of which is unique and distinguishable, thus it is the perfect implementation stand-in for the aforementioned DSOLxxxx tokens. The Ethereum platform also offers the ERC20 token, which is essentially equivalent to the AIBC system-wide unified value measure, the CFTX token.

Thus, the first release of AIBC ecosystem contains DSOLs, each with 1,000 ERC721 based unique and distinguishable DSOLxxxx. For the entire AIBC ecosystem, the unified value measure is the ERC20 based CFTX. The (fractional) ownership for each DSOL can only be transferred through auctions of its associated DSOLxxxx tokens. While theoretically all DSOLxxxx should have the same book value (each represents one thousandth of the DSOL value), since they are ERC721 tokens and each is unique and distinguishable, it is therefore possible that each DSOLxxxx has a different market value for a variety of reasons. For example, DSOL1000 might have a higher market value than the others, as it can be regarded as a “collectable item.”

## 7.4 AIBC Ecosystem Diagram

The dual-token mechanism of the AIBC ecosystem makes each DSOL an independent distributed application that is open to the public chain. This design makes AIBC an optimized asset securitization and asset anchoring platform. Figure 7.1 is a schematic diagram of the AIBC ecosystem.

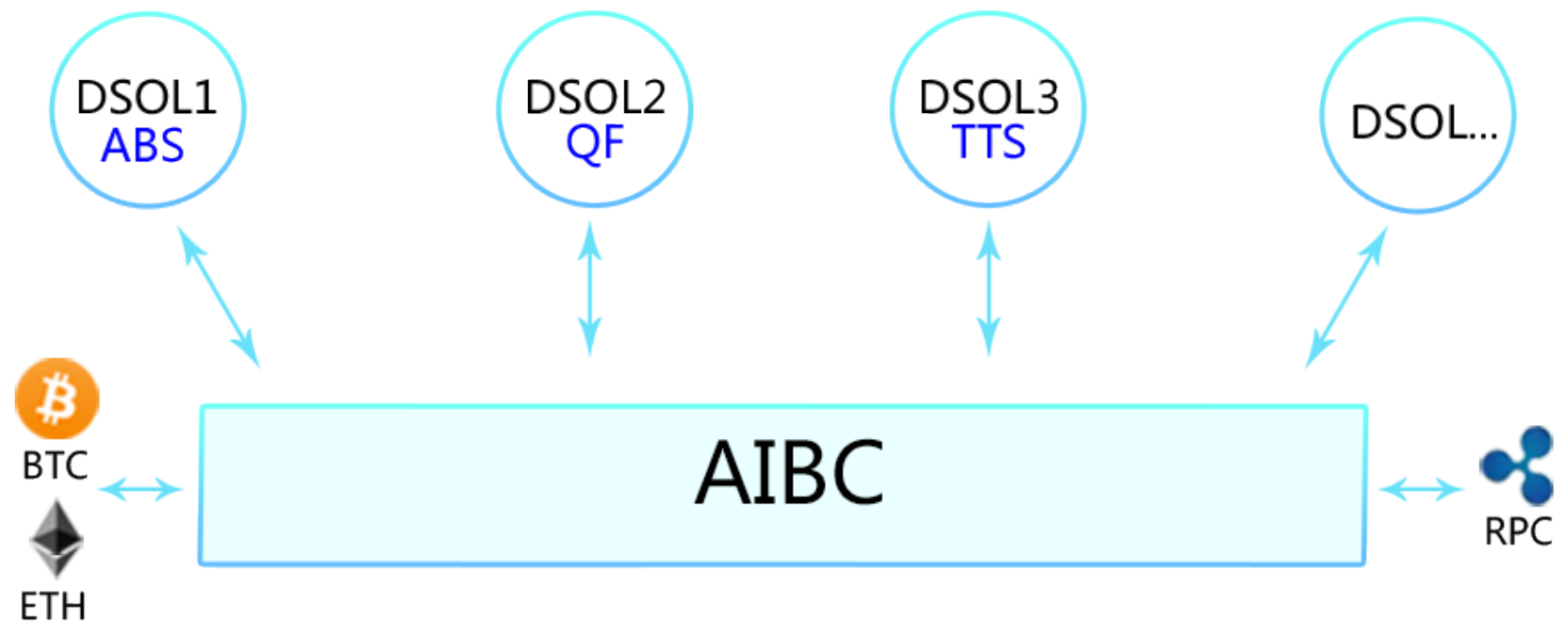

**Figure 7.1 - Schematic Diagram of the AIBC Ecosystem**

# 8 AIBC Cross-Chain Exchange Protocol Asset Securitization and Asset Anchoring

## 8.1 Asset Tokenization and Anchoring

### 8.1.1 "Stable" Digital Tokens

Currently, there are several "Stable" digital tokens available in the market. For example Tether's USDT, based on the stable value currency – US Dollar (USD). It means to anchor 1USDT=1USD, which is “the exchange rate stability token” (Tether, 2018). Tether’s definition of USDT is: "Tethers are currency-linked digital tokens, all Tethers will initially be issued on the Bitcoin blockchain via the Omni Layer protocol and so they exist as a cryptocurrency token. Each Tether unit issued into circulation is backed in a one-to-one ratio (i.e. one Tether USDT is one US dollar) by the corresponding fiat currency unit held in deposit by Hong Kong based Tether Limited. Tethers may be redeemable/exchangeable for the underlying fiat currency pursuant to Tether Limited’s terms of service or, if the holder prefers, the equivalent spot value in Bitcoin. The value of Tether is always linked to the fiat currency, and at any given time the amount of fiat currency held in our reserves will be equal to or greater than the number of tethers in circulation. From technology perspective it continues to comply with the characteristics and functions of the Bitcoin blockchain. ”

In terms of value, USDT can be analyzed from three dimensions.

1. Value Scale - Visual Display of Currency Value: USDT can directly measure the fiat value of virtual currency, and can also be regarded as the pricing of us dollar. It is particularly useful in currency transaction.
2. Circulation Medium - Intermediary for Virtual Currency Exchange: Since USDT acts as an intermediary in the circulation of virtual currency, it has become the settlement currency in digital currency and used to exchange various currencies. We can say that USDT has solved the problem of “digital token cannot be exchanged directly with fiat”.

3. Storage Medium - Safe-haven Storage: Digital token is very unstable, and in essence it's not much a currency as commodity. However, that of USDT value is stable, and the digital token retain his stored value as long as the owner hold the USDT. When the overall market is down, the holder can play a safe haven as long as the instant currency transaction exchanges other digital certificates into USDT. And because each USDT is backed up by a US dollar in Tether reserves, it can be redeemed through the Tether platform. The USDT can be used for transfer, storage, payment, etc., just like Bitcoin or any other digital token. In terms of compliance, all operations involving fiat currency require users to complete KYC certification.

The USDT is also not free of risk, at first, USDT is essentially a trust on Tether, meaning that if the company suddenly disappears or secretly makes a large number of additional issuances, then USDT will face the risk of deflation. According to Tether White Paper, "Tether is a Fiat Currency Using Bitcoin Blockchain for Transactions" further say, Tether is a decentralized digital token, but we are not a completely decentralized company and we store all the assets as a centralized pledge. The possible risks are:

1. The company may go bankrupt;
2. The bank in which the company opens account may go bankrupt;
3. The bank may freeze the funds;
4. The company may make a donation;
5. Re-centralizing the risk can paralyze the entire system.

In September 2017, Friedman LLP, a financial audit firm for tether, issued an evaluation report about tether, stated that there is no relevant assessment of the bank account published by Tether and besides tokens and currency exchange the company may involve in other investment activities. But on the evening of January 28, 2018, the partnership between Tether and the audit company Friedman LLP was officially closed, audit cannot be carried out in a sort period. This shows that the hidden risk behind USDT as the largest stable digital currency on the market cannot be underestimated.

### 8.1.2 Anchored Assets and Ownership

AIBC creatively uses blockchain technology to communicate the value of asset anchoring, such as when the value of asset changes, the digital assets on the blockchain can also be anchored accordingly, thus enable more convenient conversion of values on the Internet.

In the process of asset anchoring, what type of assets are suitable for the blockchain are explained as follows:

1. Intangible Assets: Many assets are legally referred to as “Intangible Assets”. They exist only because of the operation of law, and there is no physical entity. Intangible assets include Patents, Carbon Credits, Trademarks, Copyrights, etc. Intangible assets doesn’t involve physical entities, therefor they can be easily integrated with blockchain-based digital systems. The biggest challenge for intangible assets is to ensure that the asset transfer model in the blockchain system is in line with the legal transfer model of the real world. Differences in laws and regulations among different regions may make the asset transfer difficult (especially copyrights, where the laws vary from country to country). In summary, intangible assets are generally more easily certified than physical entities, because there is no need to worry too much about the storage and shipment of assets.

2. Interchangeable Assets: Assets are of two types exchangeable and nonexchangeable. An exchangeable asset is one that can be interchanged with another identical product such as Wheat, Gold and Water. Interchangeable assets are easier to convert into a digital certificate because they can usually be broken down into smaller units, just like bitcoin. And a certificate can represent a group of objects (i.e., a bunch of gold) rather than a group of separate objects (i.e. a warehouse filled with unique Artwork). Non-interchangeable assets must pass through an abstraction layer to be certified. For example, a company that combines assets and packages them as a whole. This is a common method of mortgages securitization, where a number of mortgages with different characteristics are combined to form a group of mortgages with roughly similar characteristics. Interchangeable assets are generally easier to certificate, because a general set of certificates is more easily associated with a common set of exchangeable asset components (i.e. 10 KG of Gold).

Secondly, the ownership of token and ownership of each part of the token need to be explained as follows:

1. There are many types of asset transfer and asset rights. Sometimes you can transfer only a part of the asset rights, such as someone transfer you the land usage rights for a limited time, not the ownership of the land. The development of asset ownership for thousands of years has spawned a variety of ownership and control rights, such as holding assets on behalf of others (trustee, trust). Here we must consider the government's jurisdiction, the type of law to be followed (common law or civil law), the type of assets and privileges to be transferred and other specific issues.
2. Some intangible assets can be licensed to millions of people at the same time, such as music copyrights. When a user buys a song from iTunes, he does not take ownership of the song (the ownership of the song has not changed), he just purchased a license to allow him to listen to music under certain conditions.
3. Therefore, blockchain token projects can generally be divided into two, the projects that involve the transfer of a part of certificate ownership such as music copyrights, and projects that involve the transfer of full rights of the certificate, such as the sale of real estate.

### 8.1.3 AIBC Assets Anchoring

AIBC creatively issues tokens (CFTX) with the anchor value of assets. Value of the most blockchain tokens (including Bitcoin) fluctuates dramatically, which is not really good for the asset certification model. The anchor value of token used in AIBC is not a hype of digital currency neither simply a digital certificate. It revitalize more asset packs, on-chain digitization and more interactive.

Unlike all credits in the USDT, which are only guaranteed by Tether, all anchors in AIBC are bound to the asset package through smart contracts. All asset packages are based on legal evidence.

First, AIBC ensures consistency between the token and reality. Traditional tokens like bitcoins will always be consistent. Every transaction follow the rules of one specific software and there are no exceptions. But in the real world, accidents often occur: gold bars are stolen, houses are burned, downloaded music go pirated and diamonds cannot be delivered properly. Because human sometimes do not follow the rules, therefore, the main challenge for any system that involve assets in the real world is to ensure the linkage of digital certificates to assets in the real world. Imagine a certificate that represents the value of all the gold bars in the vault. If a gold bar is taken from the vault, how will digital token reflect this change? Who will guarantee that the value of this digital token will be consistent to the gold bars that should have been in the vault, not the remaining gold bars in the vault? Who will bear this risk and how?

If the purchaser of the token is not sure that the token is properly linked to the assets in the real world, then the value of the token will drop and even become worthless. And because of the surety and value measurement of real assets in AIBC, the significance of anchoring the certificate becomes reliable and practical.

The above analysis shows that the biggest risk in USDT comes from its own credit endorsement. In AIBC, it doesn’t rely solely on credit endorsement of a third-party, but a value collateral through all the corresponding existing asset packages of each DSOL party and the anchor value of the token must also be a fair value with legal basis. When the value of an asset changes, the anchored token will change accordingly.

AIBC enhances system availability with a simple and elegant creation and redemption mechanism to encourage more participation. AIBC achieves this by allowing DSOL owners in the ecosystem to create and exchange anchor tokens within their DSOL environment. The anchor token in DSOL is created when the DSOL asset is mortgaged or injected, and the fiat value of the asset is 1:1 exchanged with the anchor token in the system. When the DSOL owner initiates redemption of the anchor token, the corresponding deduction is performed from the mortgage asset side and when the DSOL owner saves the anchor token, it gets an asset package of the equal value. When the value of the asset changes, the corresponding change occurs through the smart contract anchor. The number of anchor certificates created and issued must never exceed the underlying asset value. The AIBC solution provides both a secure offline approval mechanism and a flexible online approval mechanism, all controlled by smart contracts.

The application of blockchain technology in Asset-Backed Securities (ABS) scenarios has good prospects for development, specifically:

1. Since the members of the alliance share the ABS ledger data, the trust endorsement is carried out by the tamperproof blockchain system, which enhances the inter-agency trust, helps to conduct business collaboration more efficiently and transparently, and improve business efficiency.
2. Use smart contracts to implement ABS key business processes, so that ABS business processes can be effectively managed, forming a complete tracking chain, eliminate the possibility of fraud in any link, reduce the risk in the process to a certain extent, and also make the business process more automated.
3. The blockchain distributed, decentralized, peer-to-peer architecture model enables equal participation of all parties involved in the system, which facilitates the participation of heterogeneous financial institutions and reduces the risk of loss of interest due to information asymmetry.
4. The regulatory firm can join as a node and obtain real-time data of the ledger, which is beneficial for the regulatory authorities to implement rules in a timely and efficient manner, reduce intermediate links, and improve the ability of intelligent supervision.

AIBC anchoring in the asset securitization, the cash flow of all assets can generate the tokens of equivalent value recorded in the blockchain. When a change occurs on the asset side, the anchored token on the blockchain also changed immediately, and the value of the asset will be recorded every time this occurs. Risk traceability is realized through the blockchain technology of AIBC, and transactions data cannot be tampered with. Figure 8.1 shows the logical diagram of asset securitization and anchoring.

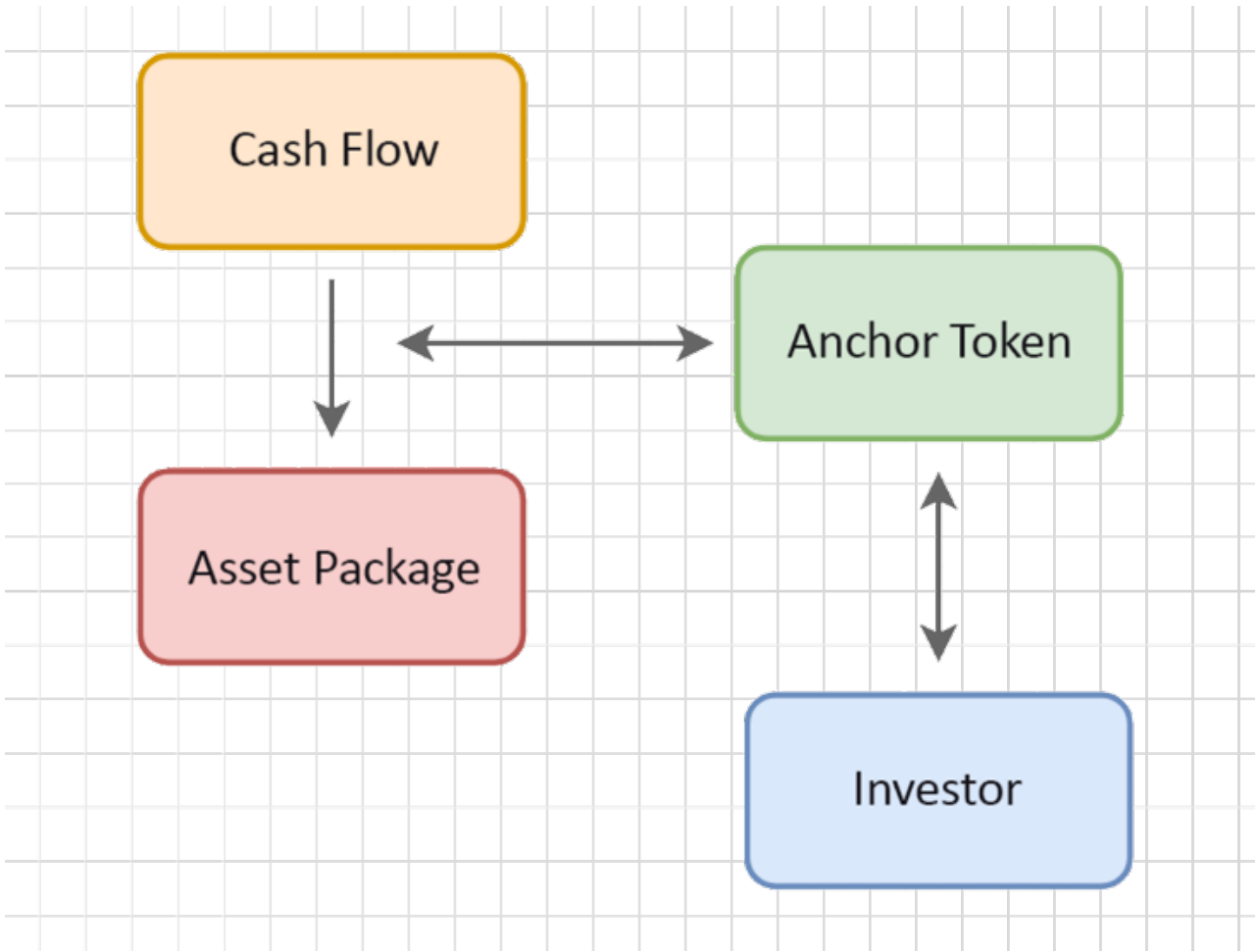

**Figure 8.1 – Logic of Asset Securitization and Anchoring**

Figure 8.2 illustrates the asset securitization and anchoring process of AIBC. The process use blockchain to realize the system supervision from three aspects: Information Disclosure, Risk Management & Control, and afterwards accountability. On one hand, the use of smart contracts to force ABS business associates to disclose relevant information in a timely and complete manner, such as asset management reports and announcements of major events etc. and issue written warning for not disclosing required information, or automatically freeze the funds in serious cases. On the other hand, the key data involved in the business process, including the cash pool situation, creditors' remittance situation, creditor credit changes and other information, are solidified and stored in real time, and traceability can be checked for any modification. Risk management is through global monitoring and risk warning through smart contracts. For example, when the total amount of market exceeds the preset threshold, the system will provide early warning and automatically prevent the generation of new special plans, and control the total amount and risk of ABS market. When the risk inevitably occurs, the blockchain traceability feature is used for analyzing the whole ABS transaction life cycle to make a complete proof chain convenient for judging the party responsible for the event and handling the corresponding liability.

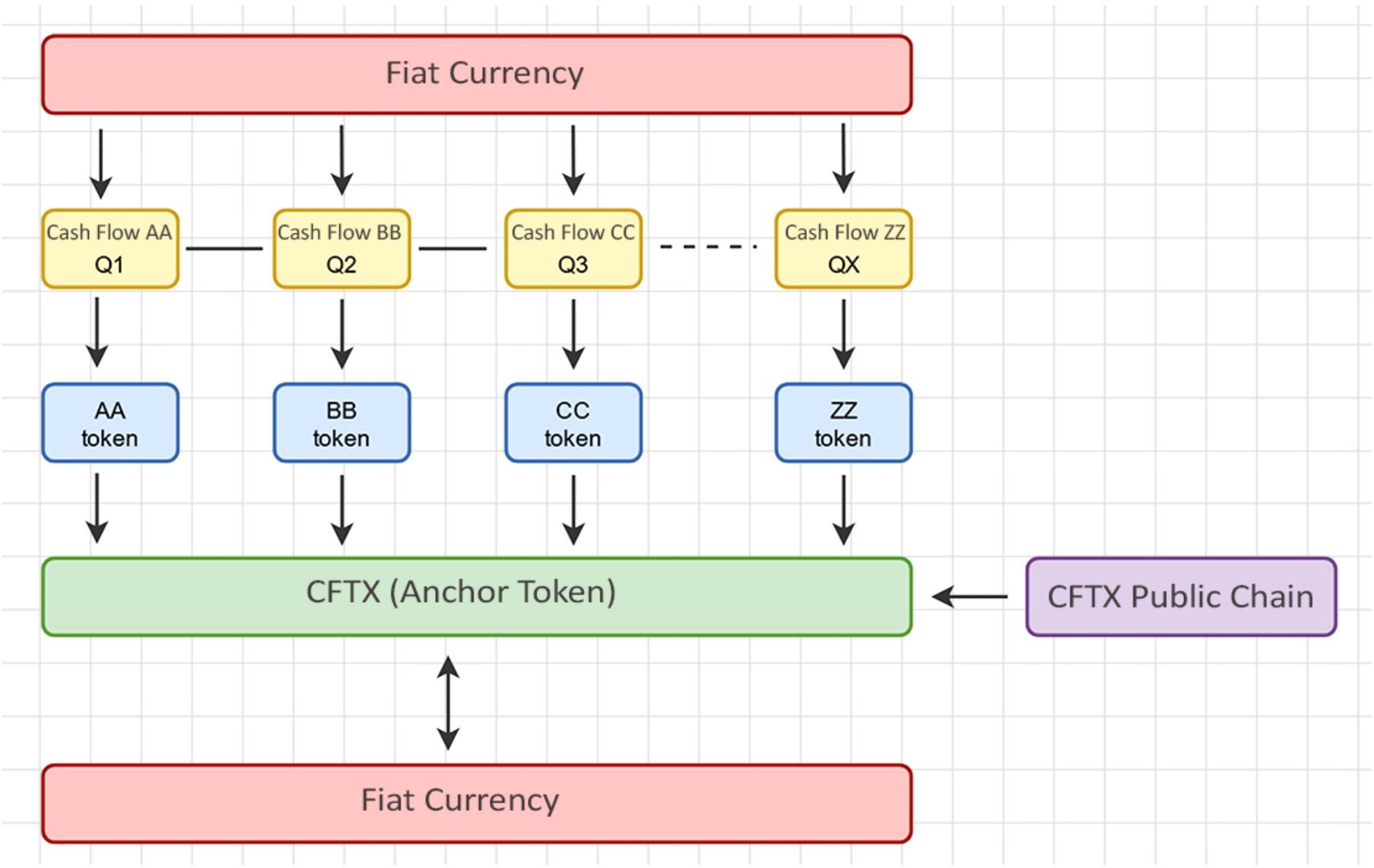

**Figure 8.2 - Asset Securitization and Anchoring Process**

## 8.2 Permission-based Cross-chain Exchange Protocol

### 8.2.1 Challenges of Cross-chain Exchange

In the AIBC ecosystem, multiple DSOLs based on the AIBC underlying public chain can be understood as independent economies. The communication between the various DSOLs and their communication with the underlying AIBC public chain (and future communication between AIBC and other public chains), must be addressed. And while having the feature of anchor tokens, AIBC is especial for sensitive information exchange.

With the continuous development of blockchain technology, the blockchain ecosystem is very much deviated from the single-chain isolated information model, forming a multifaceted environment with multiple economies and chains. New chains constantly came into existence, and the exchange between chains has become the major problem to be solved. Currently, the cross-chain approaches can be roughly divided into three categories, but they have various defects, briefly described below:

1. Notary Schemes: In the notary scheme, a trusted entity or group of entities declare to chain X that an event happened on chain Y or the event happened is correct. These groups can both automatically listen to and respond to events, as well as listen and respond on request. The notary scheme has received a lot of attention in the field of licensing, as it provides both a major competitor with a flexible consensus and no need for expensive work proofs or complex proofs of interest mechanisms. However, the shortcomings of the notary scheme are also obvious. Notary needs verification from many places. The notary is a third party and privileged institution. It can easily become the weakest link in the whole system.
2. Sidechains/Relays: If a chain B can have all the functions of another chain A, then chain B is the side chain of chain A, and chain A is the main chain of chain B. Where the main chain A is not aware of the presence of side chain B, and the side chain B is aware of chain A. Suppose blockchain has the block Header and Body. Header has the verification information such as Merkle. The header of chain A can be written into the block of chain B. Chain B uses the same consensus verification method as chain A, for example. PoW verification difficulty and length, PBFT verification voting, etc. While looking at the header of chain A, chain B can prove the data and operation of chain A from the verification information of the Merkle. Chains A and B cannot directly verify each other blocks that can create a deadlock, but the feasible way is to verify small sections at a time. This blockchain verification logic can be implemented by the chain protocol itself or by application contract. The core code needs to be exist at the same time in the two chains using the relay/sidechain scheme, but at the same time, the token or the content on both chains can be issued in unlimited amount, so that the verification process can be guaranteed without errors, thereby achieving asset transfer operations.

3. Hash-locking: The design of hash-locking scheme aims to make chain A and chain B know each other as little as possible and act as a means of eliminating the trust of notary. Hash lock originated from the Bitcoin lightning network. The lightning network itself was a means of fast payment, later its key technique hash time lock contract was applied to the cross-chain technology. Although hash locking bring about cross-chain exchange of assets, but does not realized it, and cannot achieve such cross-chain contracts, so its application scenario is relatively limited.

### 8.2.2 AIBC Permission-based Cross-chain Token-exchange Protocol

In summary, there is no clear and uniform exchange standard for cross-chain exchange. Moreover, various cross-chain methods that existed are for the token of no-collection attributes, and there is no cross-chain exchange protocol with the collection property. While AIBC adopted a standard protocol for permission-based cross-chain exchange as follows to solve the above problems:

1. AIBC creatively propose a rights-based cross-chain exchange protocol, which realize the exchange of tokens with no-collection attribute and tokens with collection attributes.
2. Through this protocol, cross-chain connection can be achieved without third party.
3. With the collection property attribute, private data can be effectively protected during the cross-chain transfer without any loss or leakage
4. After cross-chain exchange, the economic value of token will be equivalent to that before the exchange.
5. Through the implementation of this standard protocol, Information Island among chains can be completely broken, forming a real cross-chain ecosystem.

AIBC permission-based cross-chain exchange protocol implementation steps and results are as follows:

1. At the time of token creation, setup the token information basic privileges, and divide the token data into two, Private data and Public data.
2. Set the authorization data encryption method and related Key data.
3. Set the status identifier of the token data. Generally, there are three states: Normal, Exchange, and Disable where the default is Normal.
4. Suppose there are two chains A and B. The token XA in chain A is to be cross-chain exchanged with the token XB in chain B.
5. Chain A will first lock the XA token and updates the XA status flag.
6. Chain A initiates an exchange request, and the request information includes XA type, time stamp, expiration time and price etc.
7. After receiving the exchange request, Chain B will confirm the exchange request. If the exchange request is confirmed, Chain B will quickly lock the token XB and notify Chain A that the XB is locked. Otherwise, send cancellation request to chain A with the reason attached.
8. After the chain A receives the information of the token XB, the state of the XA information on the chain A is updated on time T to the disabled state, and the information that XA allows to exchange is packed and sent to chain B.
9. Chain B performs verification after receiving the information of chain A, and generates information of XA on chain B according to the information, become BXA token.
10. After the BXA token is generated and its status is set to the cross-chain exchange state, chain B packages the information of authorized exchange of token XB and send it to chain A, then performs steps 8-10 to generate the token AXB.
11. When the information exchange between chain A and chain B is completed, chain A and B initiates final confirmation message for this transaction. After receiving the confirmation message, the status of BXA and AXB is updated to “Normal” and the cross-chain exchange is completed.
12. When chain A or chain B breaks the agreement, it will be added to the blacklist. If a transaction didn’t fully execute, it’s impossible to do cross-chain exchange again.

13. Note: When the BXA is re-exchanged back to the A chain, its original authorization information can be retrieved without loss of information.

Permission-based cross-chain exchange protocol has the following advantages:

1. Through the above steps, the cross-chain exchange can be successfully completed, without the aid of third party, sidechain or other means. In the case of guaranteeing the privacy of authorized data, the cross-chain data token is completed perfectly. It supports both the exchange with non-collection attributes as well as with collection attributes.
2. Through the permission-based cross-chain token exchange standard protocol, the information island among different blockchain ecosystems is opened and connect all the blockchains as a BlockCloud.

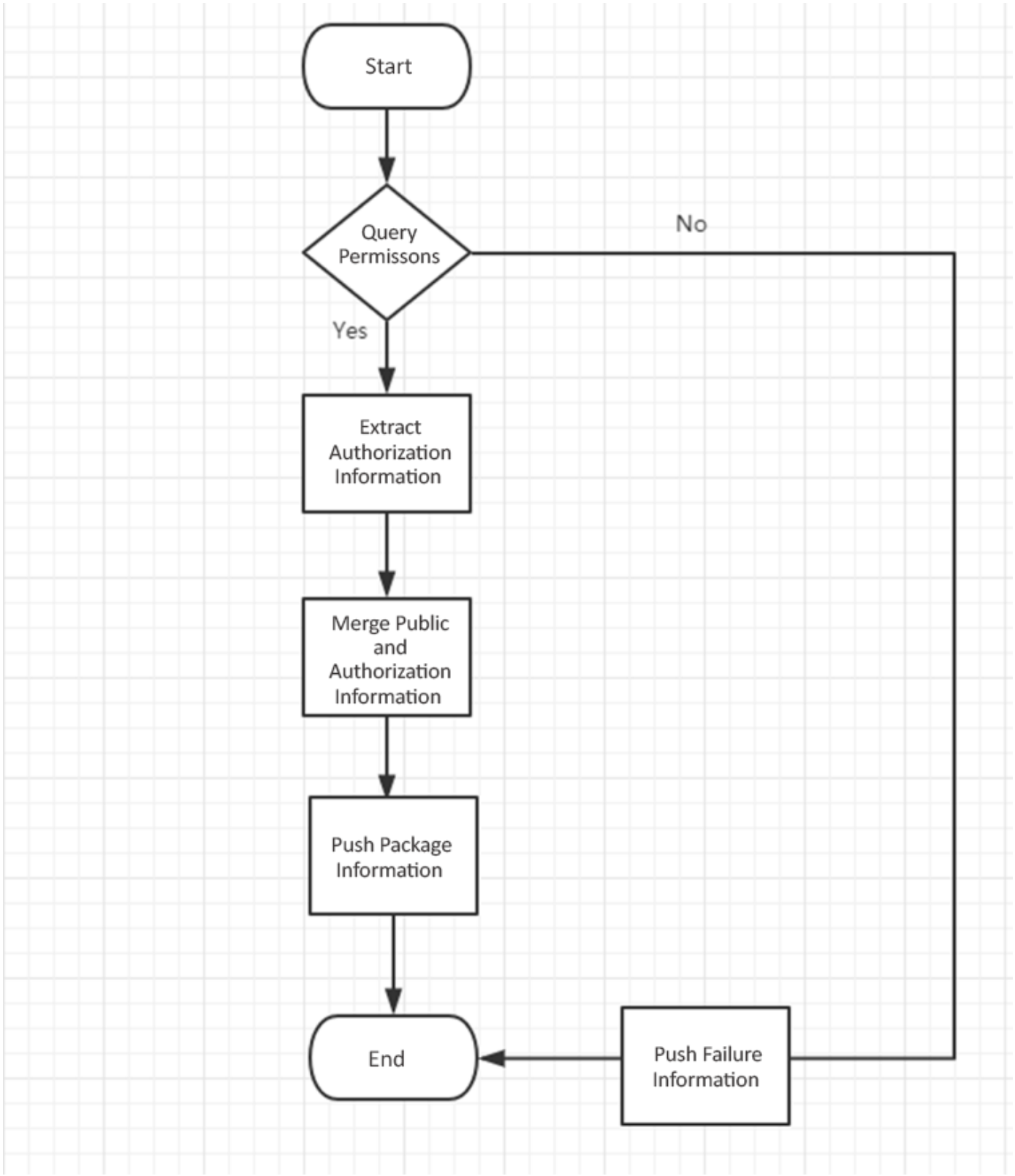

**Figure 8.3 - Authorization Information Packaging Process Flow chart**

### 8.2.3 AIBC Permission-Based Cross-chain Token-exchange Example

Implementation example is explained with images. The entire content mentioned above in the third part of technical solution is included and a detailed description is given. Each image is explained and if there are multiple implementations each one is specifically described.

We use the cross-chain exchange of ERC721 tokens on two public chains as example to describe the cross-chain information structure and then describe the cross-chain process.

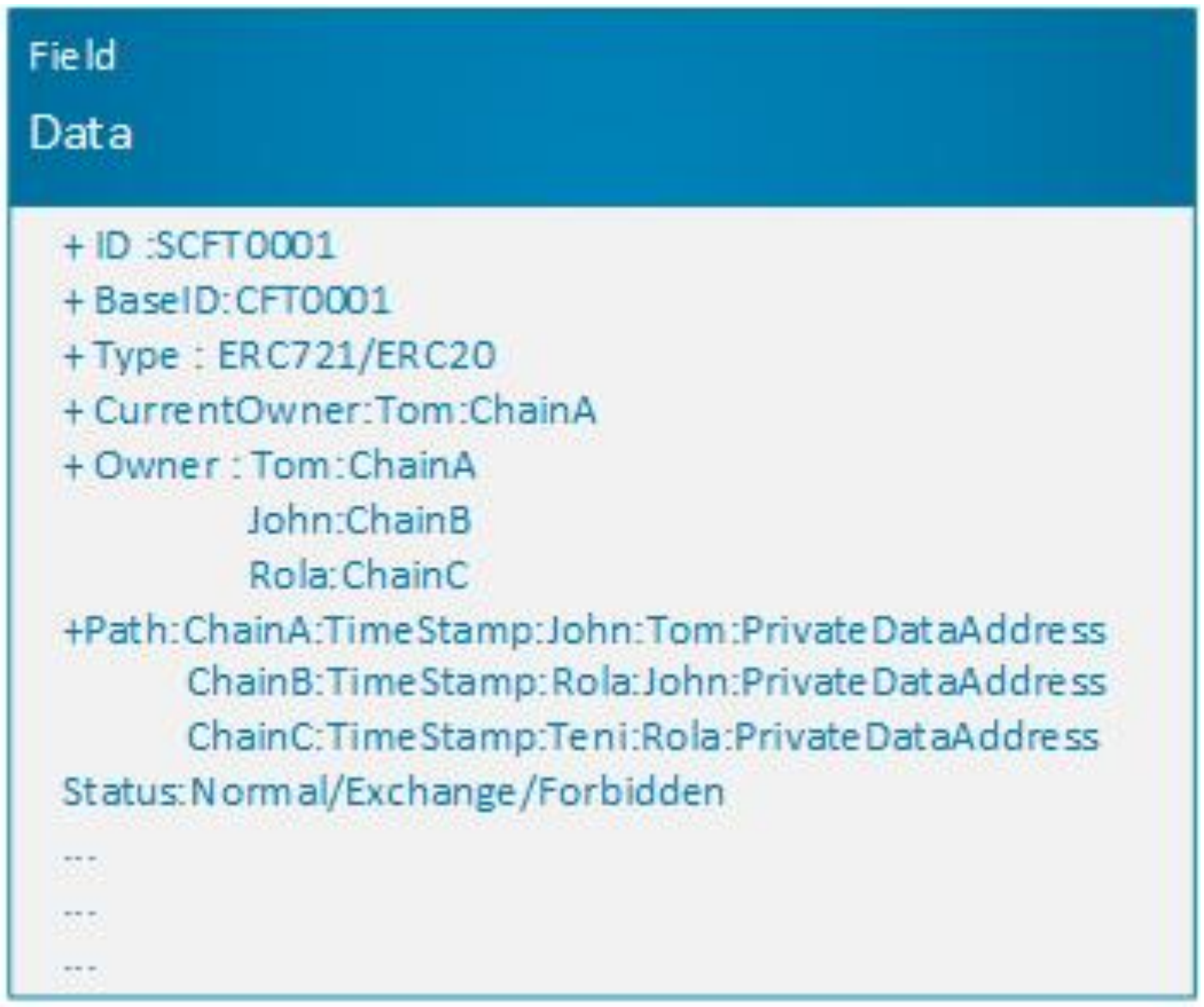

Figure 8.4 - Authorization Information Packaging Process Class Diagram

Here is a simple description of cross-chain information main attributes shown in figure 8.4

1. The “ID” is the current ID of the token that is unique in the current chain.
2. “BaseID” is the token base ID and will never change, but the ID will change when it is cross-chained. Through BaseID we can trace the overall history of the token from birth to the current state. When the token is exchanged back to the existing chain, the information of its private attribute can be restored.

3. “Type” mainly distinguishes whether the token has the asset collection attribute or not, if it has no collection, the exchange process will be simplified. But Type will not change since it was born.
4. “CurrentOwner” identifies the current owner information and current chain.
5. The “Owner” is a collection of final owners on each chain, which can effectively help the information confirmation during cross-chain recovery.
6. “Path” identifies the path information of the exchange, including all valid information as much as possible – mainly in the chain such as the exchange time, owner before, current owner, address of private attributes (if authorized, otherwise it will be empty) and so on.
7. “Status” is the status of token, that can me Normal, Exchange or Forbidden.
8. The remaining attributes will be explained later.

Suppose there are two tokens of type ERC721, a Heart token XA on chain A, and a Moon token XB on Chain B. Chain A and chain B want to exchange, but don’t want their private data to leak across the chain. The cross-chain exchange shown in Figure 8.5 can be done as given bellow:

1. Chain A loads the cross-chain exchange plugin and registers on chain B.
2. Chain B loads the cross-chain exchange plugin and registers on chain A.
3. After chain A and chain B are successfully registered in their respective routes, the cross-chain token can be exchanged.
4. The Heart token on chain A initiates a cross-chain exchange request.
5. When the Moon token on chain B receives the exchange request and agrees to exchange, it returns a consent notice to enter the cross-chain exchange process.
6. The Moon token on chain B and the Heart token on chain A respectively obtain the authorization information. When the authorization is successful, the public data and the authorization data are packaged and pushed to the opposite party. When the information is received, shadow tokens are generated on the respective public chains.

7. After the shadow token (cross-chain token) is generated, broadcast the whole network and notify the opposite party as well.
8. After both sides receive the notice, they will update their respective shadow tokens to the "Active" state and disable the original token. Thus cross-chain exchange is completed.

Through the above cross-chain exchange process, the private data of Moon and Love tokens are protected, and the authorization data can be retrieved when these tokens are exchanged back to the previous public chain. This ensures that the blockchain is open, transparent, and unchangeable.

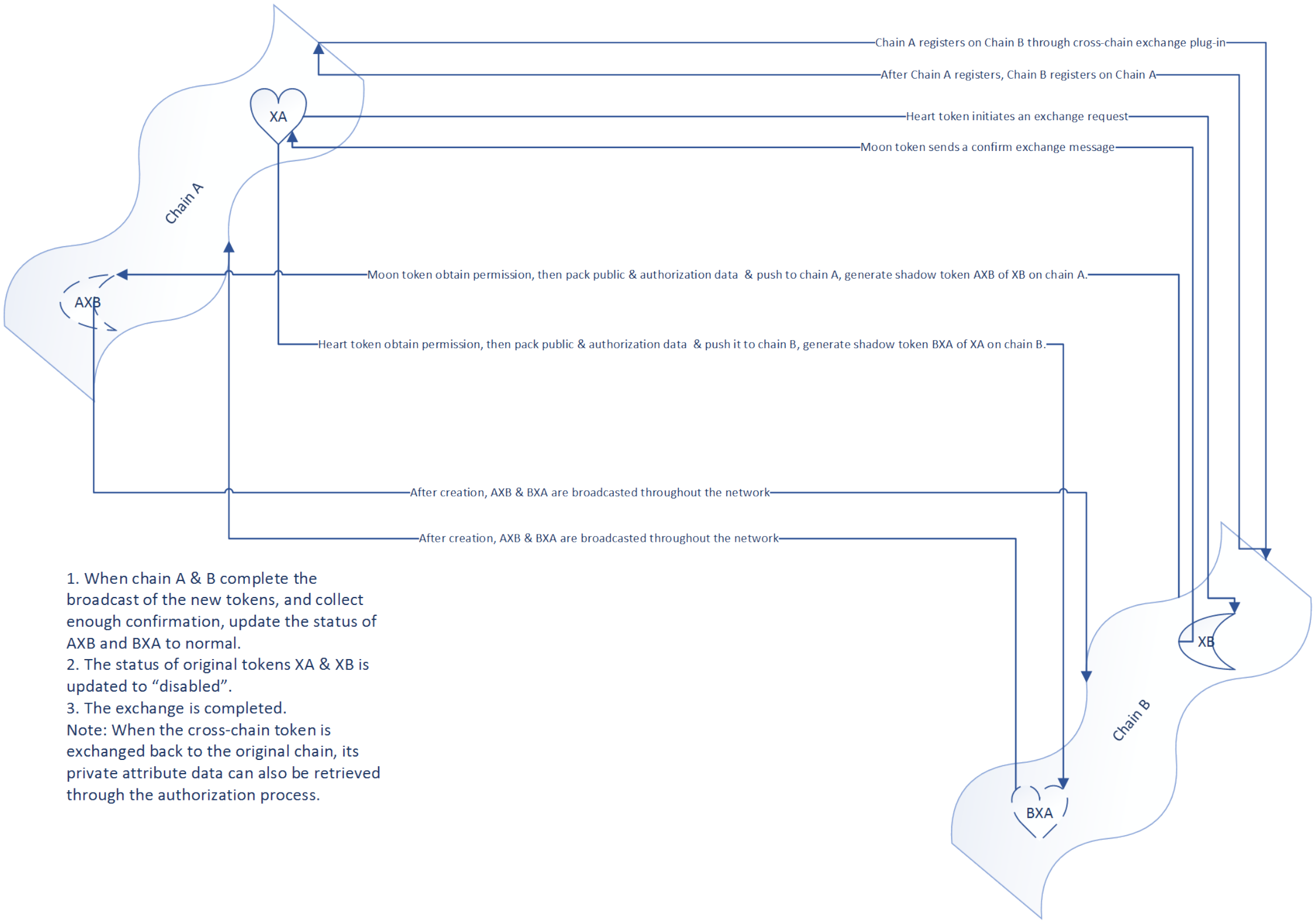

**Figure 8.5 - Authorization Information Packaging Process**

# 9 Application Scenario: Asset Securitization, Digitization and Tokenization

The first application scenario provided by AIBC is the end-to-end asset securitization, digitization, and tokenization solution provided by the AIBC Assets Digitization Center.

## 9.1 Applications of Blockchain in Assets Securitization, Digitalization and Tokenization

Asset securitization refers to the process of issuing Asset-backed Securities (ABS) on the basis of future cash flow generated by the underlying assets as reimbursement support and credit enhancement through a structured design. It is a form of trading for the issuance of tradable securities, backed by a specific portfolio of assets or a specific cash flow. According to different types of securitized assets, credit asset securitization can be divided into Mortgage-Backed Securitization (MBS), Asset-Backed Securitization (ABS) and Collateralized Debt Obligation (CDO) etc.

The basic process of finance securitization: The promoter sells the securitized assets to a SPV Special Purpose Vehicle, and SPV aggregates them into Assets Pool, then issues securities financing on the financial market based on the cash flow generated by the asset pool and finally use the cash flow generated by the asset pool to pay off the issued securities.

In the past two years, the demand for asset securitization increase day by day, and have raised some risks, including imperfect credit information system, lack of refined risk management; non-standardized asset valuation and inability to reflect status of real asset. Therefore, the introduction of blockchain technology in asset securitization aim to bring significant social benefits. As distributed ledger, blockchain has natural capacity for asset securitization. The parties involved in the project will see the underlying assets more clearly, and with the help of blockchain decentralization, reliability, immutability and trustworthiness effectively solve the problems exist in asset securities, such as multiple links, complicated processes and poor transparency of the underlying assets. At present, blockchain technology has not been able to provide end-to-end solutions in the field of asset securitization.

## 9.2 AIBC Asset Securitization, Digitization, and Tokenization Solutions

AIBC provides E2E asset Securitization, Digitization, Anchoring, and Tokenization solutions. This section gives an example solution based on actual application scenario (Commercial Real Estate Operations).

### 9.2.1 Cash Flow

One-time cash flow:

1. Property management rights for 30 years
2. Use right of supporting residential apartments for 30 years

Periodic Cash Flow:

1. Rent: Other properties (one-time cash flow)
2. Property fee: All properties

Cash flow main risks:

1. Estimation Risk: Rent miss-prediction, rent growth miss-prediction
2. Credit Risk: Prediction of rental default probability
3. Tenant Default Risk: Cash flow gap caused by tenant default
4. Tenant Bond Matching Risk: The lease does not match the underlying bond maturity

The establishment of an ad hoc fund pool:

5. Percentage of project cash flow
6. Not greater than the total circulation of CDO secondary equity layer: detailed below

The minimum indivisible cash flow calculation is based on the smallest unit of the property, for example, the monthly rental and cash flow of a property unit in order to predict, securitization, digitization and tokenization.

### 9.2.2 ABS Structured Design

The basic product categories (underlying bonds):

1. Short-term bonds (2 years): one-time property cash flow mortgage
2. Medium-term bonds (3, 5 and 7 years): short-medium tenant rent, property cash-flow mortgage
3. Long-term bonds (9, 10, 12 and 15 years): medium-long term tenant rent, property cash-flow mortgage.

Bond rating criterion:

1. Above the secondary equity level (excluding secondary equity level) contain only industry with 90% occupancy rate.
2. Industry assessment (based on market research):
   a. 100% occupancy rate (6 grades)
   b. 95% occupancy rate (12 grades)
   c. 90% occupancy rate (4 grades)
   d. 90% occupancy rate and below
1. Purpose Evaluation:
   a. Office
   b. Business
   c. Logistics
   d. Wheelhouse
   e. Catering related

Product level main risks

1. Subscription risk: Analysis method is stress testing
2. Interest rate risk: If necessary, the low priority layer can use a floating rate.
3. Secondary equity layer liquidity risk: The solution is a third-party synthetic CDO and ad hoc fund pool

The ABS design of this project mainly adopts the CDO method. And on the basis of CDO equity tranche DSOL square bottom, it introduces the convertible bond withwarrant caused by the default of the conversion index (DSOL party default fund pool bottom), and the synthetic CDO based on the CDO equity tranche (SyntheticCDO guarantees the debt certificate, no bottom, the seller bears the credit default swap, which is the counterparty risk of the CDS).The specific design idea is shown in Figure 9.1

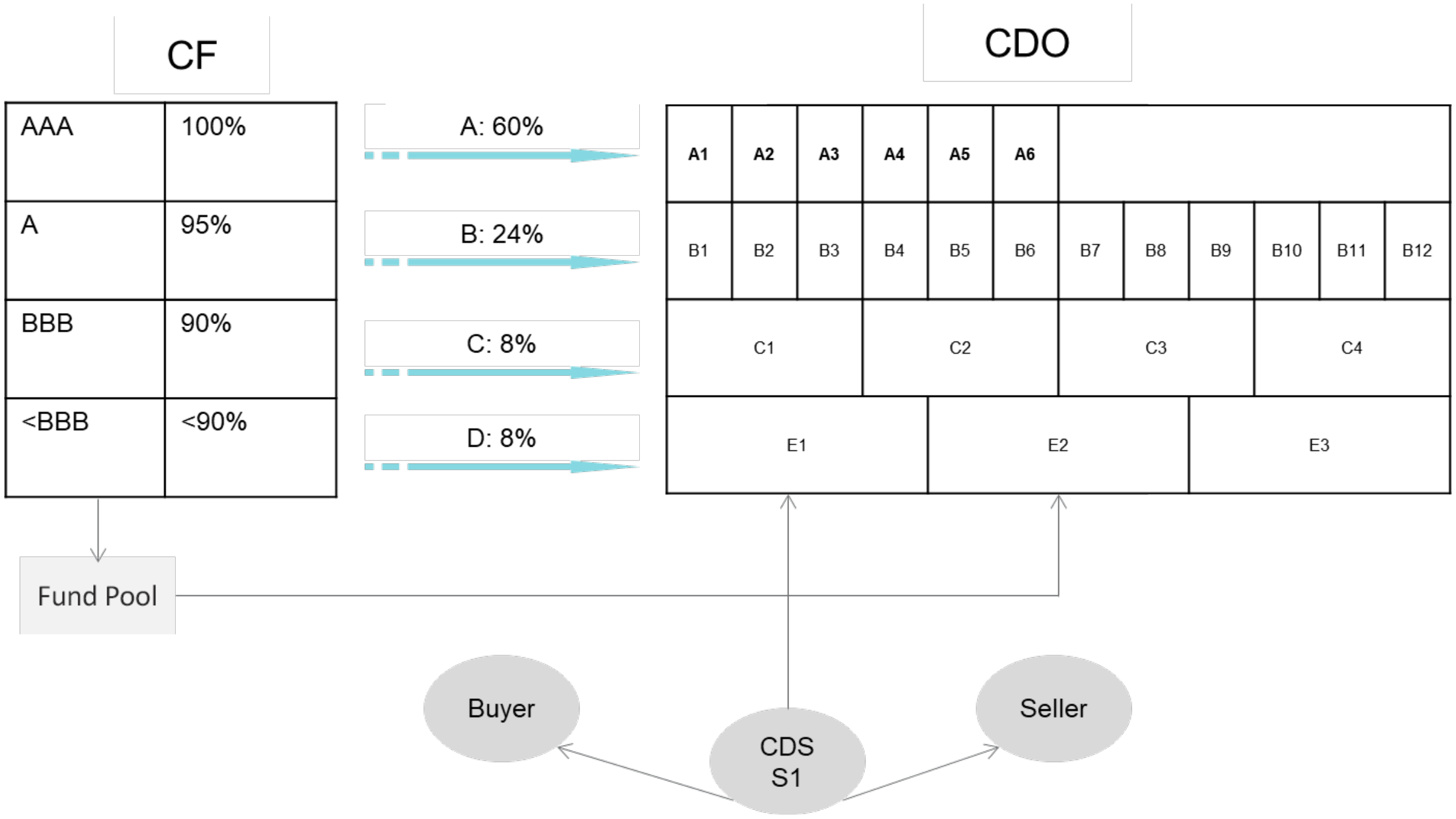

**Figure 9.1 – Asset Securitization Product Design**

For example, the total financing demand is RMB 100 million. Take 5 years lease cash flow as the asset target, 5 years bond coupon rate as the interest rate standard (4.27%) and rental rate industry assessment as the first rating standard (investors as a specific group, no compulsory third-party rating requirements). The cash-flow can be reorganized into a 5-year 4-layer structure: priority (A), intermediate (B), secondary (C), and secondary equity layer, each layer is multi-tier, and each tier is issued at a fixed interest rate, the details is given below:

1. Priority (A) layer (100% occupancy rate):
    a. Issue RMB 60 million (account for 60% of total financing)
    b. Divided into 6 tires (A1-A6), starting from the 5-year maturity coupon rate of 25bps, at interval of 5bps.

c. The coupon rates are 4.52%, 4.57%, 4.62%, 4.67%, 4.72%, and 4.77%.
   d. Pay interest every year and Principals on maturity
2. Intermediate (B) layer (occupancy rate 95%):
   a. Issue RMB 24 million (accounting for 24% of total financing)
   b. Divided into 12 tires (B1-B12), starting at 25bps above the A6 coupon rate, at 5bps interval.
   c. The coupon rates are 5.02%, 5.07%, 5.12%, 5.17%, 5.22%, 5.27%, 5.32%, 5.37%, 5.42%, 5.47%, 5.52% and 5.57%.
3. Secondary (C) layer (90% occupancy rate):
   d. Issue RMB 8 million (account for 8% of total financing)
   e. Divided into 4 tires (C1-C4), starting at 25bps above the B12 coupon rate, at 5bps interval.
   f. The coupon rates are 5.82%, 5.87%, 5.92% and 5.97%.
4. Secondary equity layer (<90% occupancy rate):
   a. Issue RMB 8 million (account for 8% of total financing)
   b. Divided into 3 tires (E1, E2, E3)
   c. The basic coupon rate of all tires is higher than that of C4 (100bps) that is 6.97%.
   a. E1 and CDO equity layer are issued simultaneously for the target synthetic CDO (S1). The E1 coupon rate is 6.97%. S1 buyer pays the purchase cost annually at a price of 25 bps, assuming that S1 secondary equity sold in full par value, the buyer pays an annual cost of RMB 20,000 (= 8 million X 0.25%). Since the S1 buyers is also an E1 investor, the overall result is that the E1/S1 investor invests 6.1% (= 6.97% - 0.25%) with the E1 of the S1 base, detail is given below:
   b. Face value of total S1 issued and face value of total E1 issued is identical, that is, how many S1s can be issued for to issue how many E1s.
   c. If the maximum purchase value of E1/S1 is equal to the total face value of the secondary equity layer (= 8 million RMB), E2 will not be issued. E2 may be issued if the maximum purchase value of E1/S1 is lower than the total face value of the secondary equity layer (<8 million RMB).

d. E2 is a convertible bond with warrant caused by a conversion indicator (default), which is different from E1.
e. The E2 investor pays the purchase cost annually at a price of 50 bps. That is, assuming that E2 is sold in full par at the subordinated equity level, the purchaser pays 40,000 (=8 million X 0.50%) of the purchase cost per year. The overall result is that the E2 investor's return is 6.47% (= 6.97% - 0.50%). See below for details.
f. If the highest purchase value of E1 and E2 is equal to the total face value of the secondary equity layer (= 8 million RMB), E3 will not be issued. E3 may be issued if the maximum purchase value of E1 and E2 is lower than the total face value of the secondary equity layer (<8 million RMB).
g. E3 is not available for sale and is owned by DSOL.

In response to the potential demand of the secondary equity investors, at the same time, issue underlying securities with the equity layer (for equity layer investors):

1. The CDO equity layer is the subject matter synthetic CDO (S1):
    a. DSOL don’t rest assure, and a third party (issuer or seller) undertakes to constitute a credit default swap, the counterparty risk of the CDS.
    b. The issuer is a third-party broker or private placement, and does not rest assure any cash-flow on DSOL.
    c. Secondary equity (E1) investors are also S1 buyers
    d. The highest purchase value of S1 is equivalent to the face value of E1, which is RMB 8 million.
    e. The S1 period is equivalent to the E1 period – 5 years
    f. The S1 buyer pays the purchase cost annually at a price of 25bps. That is, if S1 is sold in E1 par value, the buyer will pay 20,000 (=8 million X 0.25%) of the

purchase cost per year. Since the S1 buyer is also an E1 investor, eventually E1 investor invests 6.1% (= 6.97% - 0.25%) with the E1 of S1 base.

g. The default object is the secondary equity layer
h. Within the S1 term, if the E1 default occurs, the issuer guarantees that the S1 buyer will redeem the full amount of RMB, which is not more than the E1 face value of RMB 8 million.

2. Convertible bond with warrant (E2) caused by conversion indicator (default):
   a. If the maximum purchase value of E1/S1 is equal to the total face value of the secondary equity layer (= 8 million RMB), E2 will not be issued.
   b. E2 may be issued if the maximum purchase value of E1/S1 is lower than the total face value of the secondary equity layer (<8 million RMB).
   c. The maximum purchase value of E2 is equivalent to the face value of the secondary equity level, that is, RMB 8 million.
   d. E2 period is equal to E1 period – 5 years.
   e. The E2 buyer pays the purchase cost at an annual price of 50 bps. Assuming that E2 is sold in full far value of the subordinated equity, the buyer pays 40,000 (=8 million X 0.50%) of the purchase cost per year. Eventually E2 investor's return is 6.47% (= 6.97% - 0.50%).
   f. The default object is the secondary equity layer
   g. During the E2 period, if the default occurs, the E2 share is converted into the equity of ad hoc pool.
   h. The ad hoc funding pool is guaranteed by the DSOL cash-flow and regulated by the issuer. DSOL is obliged to use the surplus cash-flow to set up an ad hoc fund pool of up to the face value of secondary equity level (RMB 8 million).
   i. Due to the insufficient cash flow of DSOL, losses of E2 investors due to E2 default may not be covered by the ad hoc pool.

### 9.2.3 ABS Asset Digitization and Tokenization

This section assume Ethereum as the underlying blockchain to illustrate DSOL. The minimum investment (purchase) unit of all asset layers is RMB 10,000. We will continue the above example for explanation.

1. Priority (A) level issue RMB 60 million (account for 60% of total financing)
    a. A1-A6 level issued RMB 10 million (= 60 million / 6)
    b. A total of 1000 copies per level (= 10 million / 10,000)
    c. Each copy corresponds to one ERC721 certificate. The uniqueness of ERC721 makes each asset unique, traceable, irreversible, and makes the upstream cash flow transparent and timely reflected in the certificate.
    d. All financial attributes and actions of each ERC721 token are defined only by smart contracts.
    e. On the issue date, after each asset is purchased, its corresponding ERC721 token is activated, and each ERC721 assigned an initial 10000 DSOL private chain tokens (DSOL1).
    f. Each ERC721 generates coupon interest on a fixed date of the year. For example, if A1 interest rate is 4.52%, then 452 DSOL1 tokens will be issued on A1 anniversary. These 452 DSOL1 tokens could be transferred to the investor's specified wallet address in real time.
    g. Each ERC721 generate a coupon interest of 452 DSOL1 tokens upon maturity. The 452 DSOL1 tokens and 10,000 DSOL1 principals can be transferred to the investor's specified wallet address in real time. This ERC721 is recovered and destroyed by the DSOL.
    h. Each ERC721 reacts to the rental cash flow in real time to make it visible to the investors. This makes the ERC721 transparent and makes it easy for price trading in secondary market.
    i. The ERC721 certificate can be traded on the DSOL1 in the private chain.
2. Intermediate (B) Level (account for 95%):
    a. B1-B12 issued RMB 2 million (= 24 million / 12)

b. 200 copies per file (= 2 million / 10,000)
   c. The rest is the same as the priority (A) layer
3. Secondary (C) level (account for 90%):
   a. C1-C4 issued RMB 2 million (= 8 million / 4)
   b. 200 copies per file (= 2 million / 10,000)
   c. The rest are the same as layers A and B

The secondary equity layer is chained as follows:

1. First insist for E1 to issue RMB 8 million, but it is possible that E1 cannot issue RMB 8 million, we assume it issued RMB 4 million.
2. Based on this assumption, E1 issues 400 copies (= 4 million / 10,000), one for each E1 ERC721.
3. Synchronized with the E1 release, S1 issued 400 copies, one for each S1 ERC721.
   a. Although the issuance and subscription of E1 and S1 was synchronous, their issuers are different. E1 issuer is the DSOL side while S1 issuer is a third party. There subscribers are the same, both are E1 investors.
   b. All financial attributes and behaviors of ERC721 token are defined by the Smart Contract.
   c. On issuance date, after each E1 asset is purchased, its corresponding E1 ERC721 token is activated and an initial assignment of 10,000 DSOL private chain tokens (DSOL1) are obtained.
   d. On issuance date, at the same time each E1 is subscribed, its corresponding S1 ERC721 token is also subscribed and activated. This S1 ERC721 obtains 25 (= 10000 X 0.25%) DSOL private chain tokens (DSOL1) from the corresponding E1 ERC721, and 25 DSOL1 principals can be transferred to the investor's specified wallet address in real time.
   e. Each E1 ERC721 generates coupon interest on a fixed date of the year. The E1 coupon rate is 6.97%, resulting in 697 DSOL1 tokens on E1 anniversary.

f. At the same time, each S1 ERC721 obtains 25 DSOL private chain tokens (DSOL1) from the corresponding E1 ERC721. The 25 DSOL1 Principals can be transferred to the investor's specified wallet address in real time.
g. Eventually, each E1 ERC721 generates this profit of 672 (= 697 – 25) DSOL1 tokens and them to the investor's specified wallet address in real time.
h. If there is no default, each E1 ERC721 will yield 697 DSOL1 tokens upon maturity (5 years). The 697 DSOL1 tokens and 10,000 DSOL1 Principals can be transferred to the investor's specified wallet address in real time. The E1 ERC721 is recovered and destroyed by DSOL.
i. If there is no default, then each E1 ERC721 corresponding S1 ERC721 is retrieved and destroyed by the issuer.
j. If the default occurs, that is, the DSOL failed to generate sufficient cash flow to pay 697 DSOL1 tokens, then S1 ERC721 issuer pays S1 ERC721 purchaser equivalent to the E1 ERC721 (10,000 tokens). This E1 ERC721 is recovered and destroyed by the DSOL, and its corresponding S1 ERC721 recovered and destroyed by the issuer.
k. If the coupon interest plus principal is default at maturity, that is, DSOL cannot generate sufficient cash flow to pay 10697 (= 10000 + 697) DSOL1 tokens, then S1 ERC721 issuer pays S1 ERC721 buyer equivalent to E1 ERC721 token face value plus the interest of this period (10,697 DSOL1). These E1 ERC721 are recovered and destroyed by the DSOL and their corresponding S1 ERC721 are recovered and destroyed by the issuer. E1 ERC721 and S1 ERC721 tokens can be traded on the private chain with DSOL1.

4. Since E1 is issued 400 copies (= 4 million / 10,000), it is assumed that E2 will issue 3 million, or 300 (= 3 million / 10,000), each corresponding to one E2 ERC721 token.
    a. All financial attributes and behaviors of each ERC721 token are defined only by Smart Contracts.

b. On issuance date, after each E2 asset is purchased, its corresponding E1 ERC721 token is activated and 10,000 DSOL private chain tokens (DSOL1) are initially assigned.
c. Each E2 ERC721 generates coupon interest on the fixed annual date. The E2 coupon rate is 6.97%, generates 697 DSOL1 tokens on E2 ERC721 anniversary.
d. Simultaneously, this E2 ERC721 pays 50bps, which is 50 DSOL1, and transfers it to the DSOL specified wallet address in real time.
e. In the end, each E2 ERC721 investor’s profit is 647 (= 697 – 50) DSOL1, transferred to the investor's specified wallet address in real time.
f. If there is no default, each E2 ERC721 will generate a comprehensive income of 647 DSOL1 tokens upon maturity (5 years). The 647 DSOL1 tokens and 10,000 DSOL1 Principals can be transferred to the investor's designated wallet address in real time. This E2 ERC721 is recovered and destroyed by DSOL.
g. If the default occurs on a fixed annual date before the expiration, that is, the DSOL cannot generate enough cash flow to pay 647 DSOL1 tokens, the E2ERC721 will be converted into a 1/300 equity interest in the ad hoc pool and no more than 10,000 DSOL1 will be acquired. These E2 ERC721 are recovered and destroyed by DSOL.
h. If the coupon interest plus Principals default on maturity, that is, DSOL cannot generate enough cash flow to pay 10647 (= 10000 + 647) DSOL1 tokens, then E2 ERC721 will be converted into a 1/300 equity interest in ad hoc pool and no more than 10,647 DSOL1 will be acquired. These E2 ERC721 are recovered and destroyed by DSOL.
i. Due to the insufficient cash flow of DSOL, the losses of E2 ERC721 investors due to E2 default may not be covered by ad hoc pool.
j. E2 ERC721 tokens can be traded on the private chain with DSOL1

5. Since E1+E2 issued a total of 7 million or 700 copies (= 4 million/10,000), E3 should issue 1 million or 100 copies (= 1 million/10,000), one for each E3 ERC721 token.
   a. E3 ERC721 is not issued to investors and is owned by DSOL.

Figure 9.2 shows the ABS asset Digitization/Tokenization flow chart.

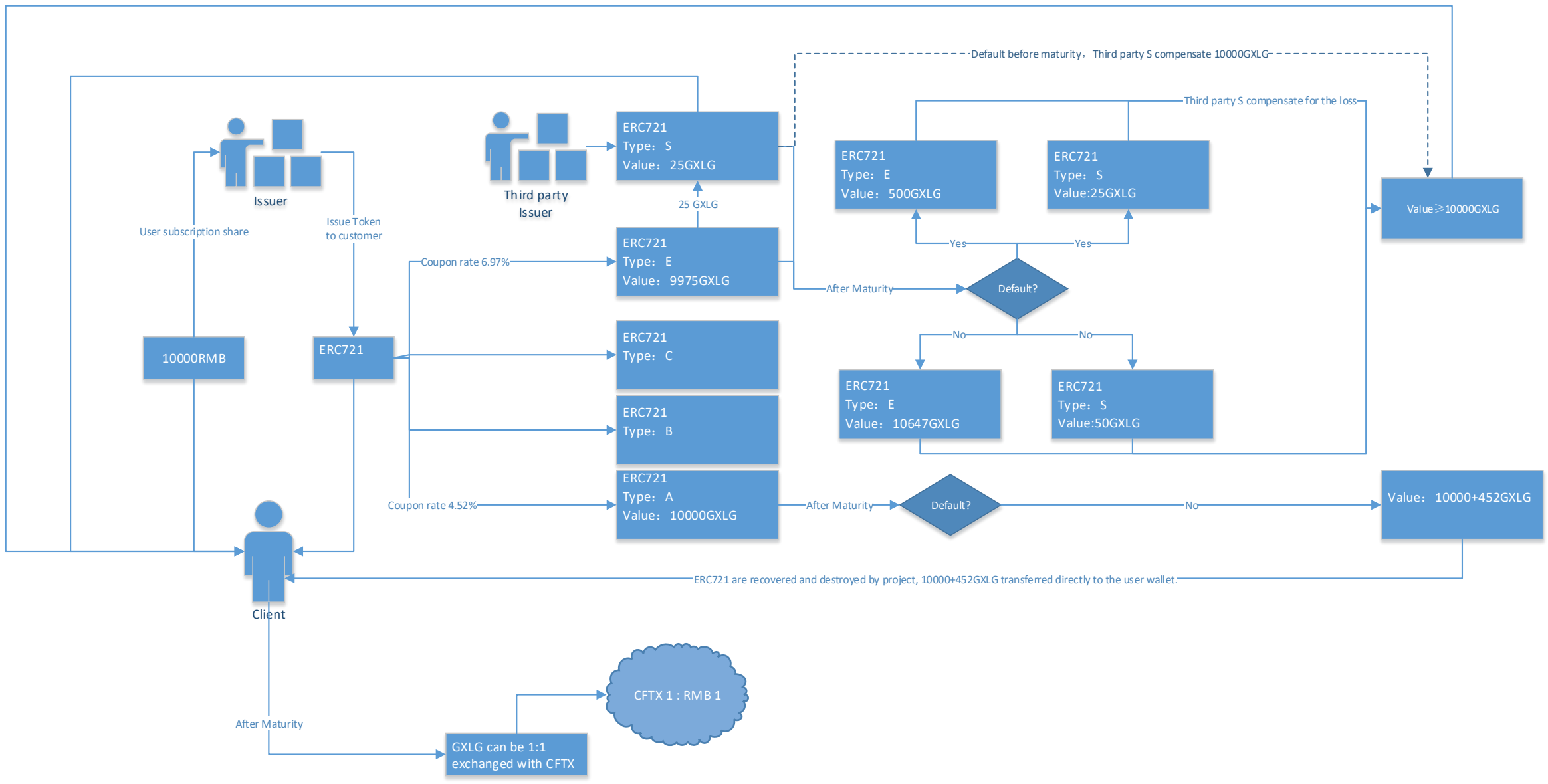

**Figure 9.2 – ABS Asset Digitization/Tokenization Flow Chart**

### 9.2.4 Business Model

A, B, C, and E level structured products are issued by DSOL through private placements specified by DSOL. The S1 level structured products are issued by private placement specified by the DSOL.

The products are sold to qualified investors by private placements. After the qualified investors subscribe, A, B, C, E, and S1 level ERC721 tokens are issued by private chain of the project. A flow chart of ABS asset Digitization/Tokenization business model is given in Figure 9.2.

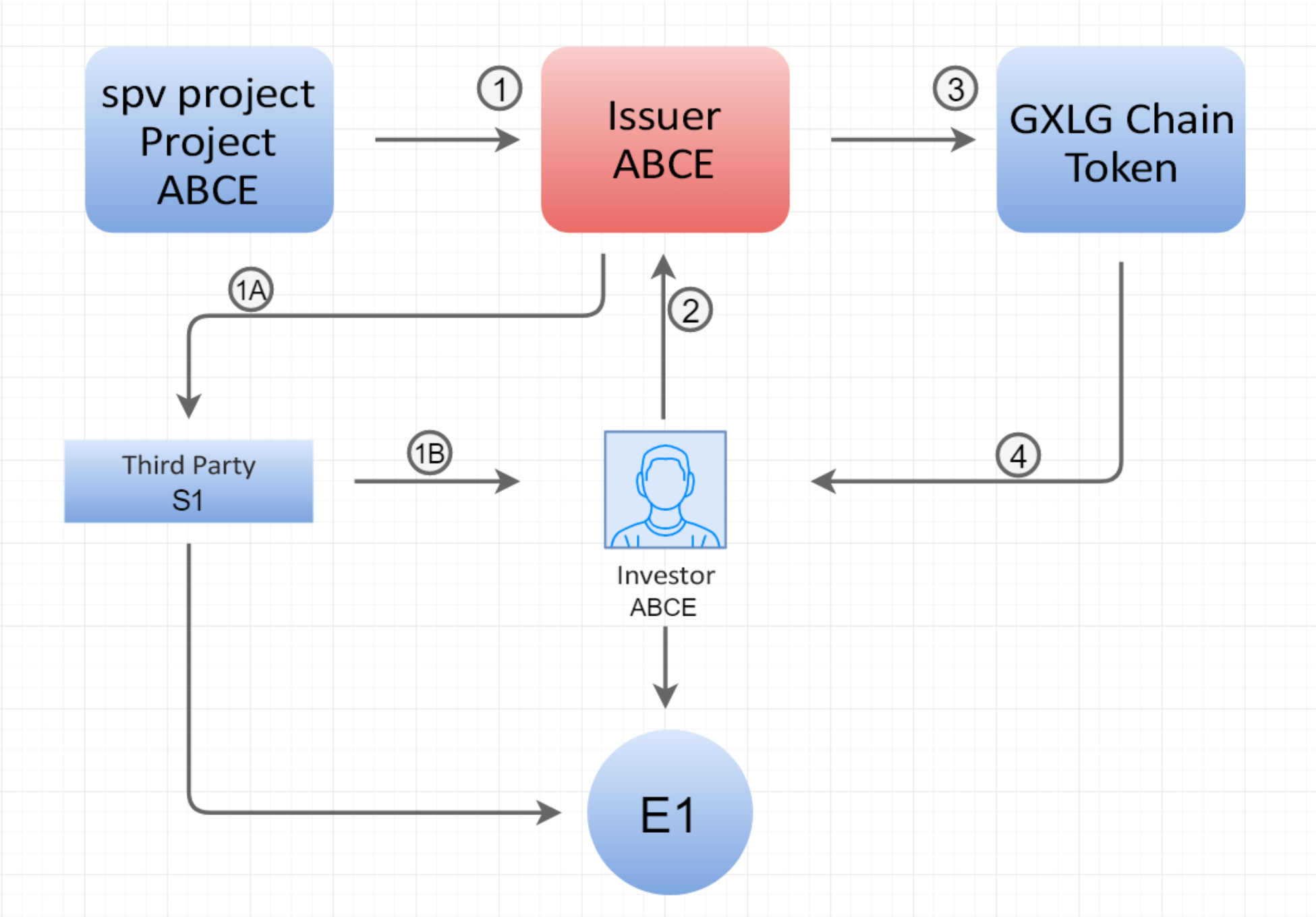

**Figure 9.3 – ABS Asset Digitization/Tokenization Business Model**

# 10 Conclusions

The AIBC is an Artificial Intelligence and blockchain technology based large-scale decentralized ecosystem that allows system-wide low-cost sharing of computing and storage resources. The AIBC consists of four layers: a fundamental layer, a resource layer, an application layer, and an ecosystem layer.

The AIBC implements a two-consensus scheme to enforce upper-layer economic policies and achieve fundamental layer performance and robustness: the DPoEV incentive consensus on the application and resource layers, and the DABFT distributed consensus on the fundamental layer. The DABFT uses deep learning techniques to predict and select the most suitable BFT algorithm in order to achieve the best balance of performance, robustness, and security. The DPoEV uses the knowledge map algorithm to accurately assess the economic value of digital assets.

The AIBC is task-driven with a “blocks track task” dynamic sharding structure. It is a 2D BlockCloud with side chains originated from the super nodes that track the progress of tasks. The 2D implementation makes it extremely efficient to evaluate the incremental economic value of additional knowledge contributed by each task. The dynamic sharding feature resolves the scalability issue and improves the efficiency further.

The AIBC has a dual-token implementation. In addition to the system-wide unified measure of value and transaction medium CFTX, each DSOL will have a separate numbering interval as a single distinguishable token, DSOLxxxx. The dual-token approach allows the CFTX be used for the entire AIBC ecosystem, while enables the transfer of DSOL ownership through auctions of DSOLxxxx tokens.

Based on the dual-token platform, AIBC creatively issue the token with asset anchoring value (CFTX). The anchor token in DSOL is created when the DSOL asset is mortgaged or injected, and the currency denominated assets are 1:1 exchanged into the system to anchor the certificate. When the value of the asset changes, the corresponding change occurs through the smart contract anchor. The communication between various DSOLs and their communication with the underlying AIBC public chain (and future communication between AIBC and other public chains), especially sensitive information, is done by AIBC's standard protocol for cross-chain exchange

The first application scenario that AIBC will provide is the asset Securitization and Tokenization. The Cofintelligence BlockCloud Assets Digitization Center has developed a decentralized channel for digital assets trading, effectively transforming these digital assets transactions into social communication networks, increase the flow of assets and thus increase their value.

The second application scenario that the AIBC will serve is the robotic investment advisor. The Cofintelligence BlockCloud AI research team has developed innovative and advanced ML and AI models for quantitative investment that have consistently exceeded investors' expectations.

The AIBC combines Artificial Intelligence, Big Data, Cloud Computing, and Distributed Databases to provide platforms and algorithms for applications in Smart Investment, Asset Management, Asset Digitization, Supply Chain Finance, BlockCloud services and many other emerging areas.